\documentclass[twocolumn]{article}
\usepackage[utf8]{inputenc}
\usepackage{amsmath,amssymb,amsfonts}
\usepackage{float}
\usepackage{algorithm}
\usepackage{algcompatible}
\usepackage[title]{appendix}
\usepackage{graphicx}
\usepackage{hyperref}
\usepackage{textcomp}
\usepackage{xcolor}
\usepackage{url}
\usepackage{booktabs}
\usepackage{multirow}
\usepackage{threeparttable}
\usepackage{float}
\usepackage{subcaption}
\usepackage{cleveref}
\usepackage{tikz}
\usepackage{array}
\usepackage{makecell} 
\usepackage{dblfloatfix}
\usetikzlibrary{arrows.meta, positioning}

\title{State Propagation Also Satisfies: \\
A Complex-Valued State-Space Model for Deterministic State Tracking}

\author{
  Li Xiaohe$^{1}$\thanks{First author.} \\
  \small{$^1$GuangDong Police College}
  \thanks{Code and models are available at \url{https://github.com/hilhert/CSP}.}
  \and
  Yang Lu$^{2}$\thanks{Corresponding author.} \\
  \small{$^2$Department of Computer Science and Technology, School of Informatics, Xiamen University}
}

\date{}
\begin{document}
\maketitle

\begin{abstract}
Despite the dominance of massive language models, leading paradigms like Transformers and Mamba fundamentally falter at continuous deterministic state tracking, suffering from catastrophic out-of-distribution (OOD) collapse when generalizing to extended sequences. To shatter this bottleneck, we present the \textbf{Complex State Propagator (CSP)}, a radically minimalist recurrent paradigm that operates strictly on the complex phase manifold without intermediate output projections, amplitude modulations, or per-step non-linearities. Crucially, we unveil an unprecedented architectural marvel: \textbf{the phase signal survives extreme depth with absolute zero informational attenuation}. By implementing an exact four-quadrant \textbf{Coordinate-to-Phase (C-to-2)} transformation governed by the \(\text{atan2}(y, x)\) activation, CSP forces the continuous optimization landscape to seamlessly align with discrete cyclic groups. Remarkably, with a mere \textbf{3-layer hidden topology} trained on short inputs (length 16),CSP demonstrates absolute mathematical purity, achieving a nearly 
$100\%$ validation accuracy and F1-score when generalized to a 
$4\times$ prolonged OOD length of 64 on the canonical Mod-3 tracking task. Our work establishes complex-valued, pure-phase propagation not merely as a compact alternative, but as a dominant frontier that beats heavy networks at a fraction of their size.
\end{abstract}

\section{Introduction}
\subsection{Rethinking Semantic Learning and RoPE Semantics}
Sequence modeling has been reshaped by two architectural families: 
attention-based Transformers \cite{vaswani2017attention} and state-space 
models such as S4 \cite{gu2022efficiently}, S5 \cite{smith2023simplified}, 
H3 \cite{fu2023hungry}, and Mamba \cite{gu2024mamba}. These architectures 
have become the backbone of modern language models, achieving strong 
results on tasks that require learning complex semantics from data: 
translation \cite{vaswani2017attention,devlin2019bert}, language 
understanding \cite{brown2020language}, long-context reasoning 
\cite{dao2024transformers}, and multimodal generation \cite{gu2024mamba}. 
Their success has driven a broad consensus that scaling these architectures 
with more parameters and more data will continue to unlock increasingly 
general capabilities.

Yet these same architectures exhibit a striking fragility on tasks that 
are, in a precise computational sense, far simpler. Parity checking---deciding 
whether a binary sequence contains an odd number of ones---can be solved by 
a two-state automaton. Modular counting requires only a finite-state machine 
with $k$ states. Parenthesis matching is recognizable by a pushdown automaton 
with a single stack symbol. These are among the simplest problems in formal 
language theory, well within the reach of any finite-state or context-free 
computation. But recent theoretical work has established that Transformers 
with Lipschitz-continuous attention cannot solve parity, $\mathrm{Dyck}_1$, 
or $\mathrm{Dyck}_2$ for arbitrary input lengths \cite{hahn2020theoretical}. 
Even with layer normalization and task-specific positional encodings, 
empirical studies find that Transformers struggle to learn parity in 
practice \cite{bhattamishra2020ability,chiang2022overcoming}, often 
converging to low-sensitivity functions that fail to generalize beyond 
the training length \cite{anil2022exploring,bhattamishra2023transformers}. 
Similar limitations have been established for state-space models: linear 
RNNs with input-independent transitions cannot track state regardless of 
depth \cite{merrill2024illusion}, and Mamba's diagonal, non-negative 
transition matrix prevents it from solving parity in finite precision 
\cite{sarrof2024space,grazzi2025unlocking,khavari2025parity}. For modular 
counting tasks with $k > 2$, complex eigenvalues or non-diagonal structure 
are necessary \cite{grazzi2025unlocking,lumbroso2024provable}. Recent work 
has further shown that even multilayer combinations of input-dependent and 
input-independent SSM layers fail to solve parity \cite{khavari2025parity}.

The contrast is stark. A vanilla RNN with $\tanh$ activation and a single 
hidden layer can solve parity exactly, because the saturating nonlinearity 
implements a hard binary state transition 
\cite{merrill2024illusion,sarrof2024space}. A complex-valued RNN with 
learned phase rotations can track arbitrary modular counters by accumulating 
phase on the unit circle \cite{arjovsky2016unitary,lumbroso2024provable}. 
These architectures have far fewer parameters and far less expressive 
capacity than modern language models in the semantic domain, yet they 
succeed where Transformers and Mamba fail. The theoretical literature has 
begun to explain \emph{why} this gap exists---the geometry of the state 
transition matrix determines whether a model can represent discrete state 
\cite{grazzi2025unlocking,merrill2024illusion}---but the practical 
implications remain underexplored. If the architectures that dominate 
language modeling have a fundamental blind spot on state-tracking tasks, 
then understanding and closing this gap may unlock improvements in the 
domains where these models are deployed.

This paper is motivated by that possibility. We propose the \textbf{Complex 
State Propagator (CSP)}, a minimal recurrent architecture designed to 
isolate and address the state-tracking limitation directly. CSP maintains 
a complex-valued hidden state and updates it by input-dependent rotations 
in the complex plane. Each block applies an element-wise complex 
normalization that projects the state back onto the unit circle, so that 
information is carried primarily in the phase rather than the magnitude. 
No output projection is applied between blocks; the hidden state itself 
propagates through the stack. This design is motivated by the observation 
that exact state tracking requires transitions with magnitude exactly 
$1$---rotations, not decay \cite{grazzi2025unlocking,lumbroso2024provable}. 
A state transition that multiplies by a real scalar less than $1$ 
introduces leakage, and over long sequences the accumulated signal decays 
to zero. A rotation with unit magnitude preserves information indefinitely.

A second motivation comes from recent developments in positional encoding. 
Modern language models predominantly use Rotary Position Embedding (RoPE) 
\cite{su2024roformer}, which rotates query and key vectors by angles 
proportional to their absolute positions \cite{roformer2024,peng2024yarn}. 
RoPE has been shown to improve long-context generalization 
\cite{chen2023extending,peng2024yarn} and to induce specific computational 
structures within attention heads \cite{heo2024rotary}. 
But RoPE is fundamentally a \emph{positional} mechanism: it encodes where 
a token is, not what the sequence contains. The rotation angle depends only 
on the position index, not on the input content. CSP asks a different 
question: what if rotation is applied not to position, but to 
\emph{content}? What if the rotation angle at each step is a function of 
the input, so that the phase accumulates according to the sequence's 
semantic structure rather than its position index? This reframing turns 
rotation from a positional prior into a state-tracking mechanism. The gate 
that determines the rotation angle becomes the mechanism by which the model 
decides how to update its discrete state---whether to flip a parity bit, 
advance a modular counter, or push or pop a stack symbol. We believe this 
connection between rotation-based position encoding and rotation-based 
state tracking is underexplored and may offer a principled route to 
improving state-tracking ability in sequence models.

We evaluate CSP on three state-tracking tasks and compare it against five 
baselines under a standardized input pipeline. Our contributions are:

\begin{enumerate}
    \item \textbf{A three-tier empirical characterization of length 
    generalization.} Across parity, mod-3 counting, and Dyck-1 parenthesis 
    matching, we identify three tiers of behavior: exact-transition models 
    (vanilla RNN, complex RNN) that generalize perfectly; soft-contraction 
    models (CSP, Mamba, Transformer) that decay with length; and 
    attention-based models that fail on parity and mod-3 but excel on 
    Dyck-1.

    \item \textbf{CSP as the strongest soft-contraction model on state 
    tracking.} CSP-Vanilla outperforms both Mamba and the decoder-only 
    Transformer on parity and mod-3 length generalization by 5--15 
    percentage points across five random seeds. CSP-Fast, a parallel 
    formulation, shows a smaller but still positive advantage.

    \item \textbf{A complex-valued state mechanism with stable training.} 
    CSP's rotation and complex normalization maintain a bounded state 
    without gradient explosion, unlike negative-eigenvalue SSM variants 
    that require careful eigenvalue clipping.

    \item \textbf{An open-source implementation} with reproducible seeds, 
    training scripts, and evaluation protocols, available at 
    \url{https://github.com/hilhert/CSP}.
\end{enumerate}

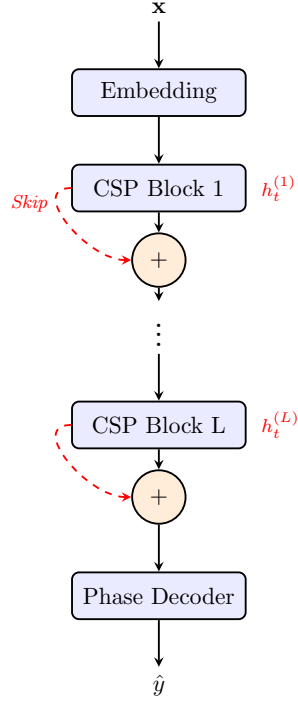
\begin{figure}[t]
\centering
\resizebox{0.5\columnwidth}{!}{%
\begin{tikzpicture}[
    node distance=0.7cm,
    block/.style={rectangle, draw=black, thick, minimum width=2.6cm, minimum height=0.7cm, align=center, rounded corners=3pt, fill=blue!8},
    addnorm/.style={circle, draw=black, thick, minimum size=0.8cm, align=center, fill=orange!15, inner sep=0pt},
    arrow/.style={->, >=stealth, thick},
    skiparrow/.style={->, >=stealth, thick, dashed, red}
]

\node (input) {$\mathbf{x}$};
\node[block, below=of input] (embed) {Embedding};

\node[block, below=of embed] (block1) {CSP Block 1};
\node[addnorm, below=0.3cm of block1] (add1) {$+$};

\node[font=\Large, below=0.2cm of add1] (dots) {$\vdots$};

\node[block, below=of dots] (blockL) {CSP Block L};
\node[addnorm, below=0.3cm of blockL] (addL) {$+$};

\node[block, below=of addL] (decoder) {Phase Decoder};
\node[below=of decoder] (output) {$\hat{y}$};

\draw[arrow] (input) -- (embed);
\draw[arrow] (embed) -- (block1);
\draw[arrow] (block1) -- (add1);
\draw[arrow] (add1) -- (dots);
\draw[arrow] (dots) -- (blockL);
\draw[arrow] (blockL) -- (addL);
\draw[arrow] (addL) -- (decoder);
\draw[arrow] (decoder) -- (output);

\draw[skiparrow] (block1.west) to[out=180, in=180, looseness=1.2] node[left, font=\footnotesize\itshape, text=red, pos=0.3] {Skip} (add1.west);
\draw[skiparrow] (blockL.west) to[out=180, in=180, looseness=1.2] node[left, font=\footnotesize\itshape, text=red, pos=0.3] {} (addL.west);

\node[right=0.1cm of block1, font=\footnotesize, text=red] {$h^{(1)}_t$};
\node[right=0.1cm of blockL, font=\footnotesize, text=red] {$h^{(L)}_t$};

\end{tikzpicture}%
}
\caption{CSP architecture. Input passes through stacked blocks; hidden states propagate forward; skip connections bypass blocks to Add \& Norm.}
\label{fig:csp_arch}
\end{figure}

\subsection{Our Contribution}

We propose the \textbf{Complex State Propagator (CSP)}, a minimalistic architecture designed \textit{from the ground up} for state tracking:
\begin{enumerate}
  
    \item \textbf{Complex-Valued States}: The state is a complex vector, updated via \textbf{learned rotations}---enabling exact phase-based discrete state transitions.
    \item \textbf{Block-level Skip Connections}: To stabilize training in deeper stacks while preserving clean state transitions, we incorporate block-level skip connections along with terminal complex-domain normalization and SiLU activation.
    \item \textbf{Phase-Focused Representation}: Information is encoded primarily in the angular dynamics of a complex-valued state, supported by complex normalization and phase-aware decoding.
    \item \textbf{Open-source implementation}: Code available at \url{https://github.com/hilhert/CSP}.
\end{enumerate}

\newpage
\section{Methodology}

\subsection{Complex State Space and Phase Representation}
We adopt a complex-valued state space because it naturally supports periodic 
representations. After element-wise complex normalization, the magnitude of 
each state is constrained to unity, so that all information is encoded 
exclusively in the phase. In this sense, the normalized complex domain reduces 
to a pure phase representation, and the central design principle of CSP 
becomes clear: the architecture is fundamentally about propagating and 
modulating phase signals across layers.

A natural starting point is to write each state as $\mathbf{h}_t = e^{i 
\boldsymbol{\phi}_t}$, where $\boldsymbol{\phi}_t \in \mathbb{R}^d$ is a 
phase vector of the same dimension as the state. This phase-only view is 
conceptually clean, but it is not directly convenient for sequential 
computation: phase alone does not admit straightforward linear transformations 
or accumulations, which are the basic building blocks of recurrence and 
attention. We therefore unfold the complex domain into two real channels, 
writing each state as $\mathbf{h}_t = [\cos \boldsymbol{\phi}_t; \sin 
\boldsymbol{\phi}_t]$, so that phase signals can be processed using standard 
real-valued machinery, such as RNN-style recurrences, Mamba-style state 
updates, and even Transformer-style blocks. The complex representation is 
preserved as the underlying semantics, while the computation itself is carried 
out in the real domain.

With this unfolding in place, we construct the complex-valued state recurrence 
as following, taken $\mathbf{z}_t$ as the recurrent input at time step t:
\begin{equation}
    \tilde{\mathbf{h}}_t = \alpha_t \mathbf{h}_{t-1} 
                         + \gamma_t \, \mathbf{B} 
                         \bigl(\mathbf{r}_t \odot \mathbf{z}_t\bigr)=\boldsymbol{\rho} e^{i\boldsymbol{\phi}_t},
\end{equation}
where $\boldsymbol{\rho}_t = |\mathbf{h}_t| \in \mathbb{R}^d$ is the 
magnitude vector and $\boldsymbol{\phi}_t \in \mathbb{R}^d$ is the phase 
vector, both of the same dimension as $\mathbf{h}_t$ and $\mathbf{z}_t$ and $\odot$ denotes element-wise multiplication. Here 
the rotation $\mathbf{r}_t = e^{i\theta_t} \mathbf{1}$ with $\theta_t \in \mathbb{R}$ a 
learnable scalar shared across all dimensions and $\mathbf{1} \in \mathbb{R}^d$ a d-dimensional all one vector, acting as an interference 
transformation between the real and imaginary components of the state. And $\mathbf{B} \in \mathbb{R}^{d \times d}$ is the state transfer matrix which diffuses state update to each dimension, it is multiplied to the real part and the image part of $\bigl(\mathbf{r}_t \odot \mathbf{z}_t\bigr)$ separately. We 
then normalize the magnitude $\boldsymbol{\rho}_t$ at each step, leaving 
the phase $\boldsymbol{\phi}_t$ as the only degree of freedom that survives 
between layers, so that the state trajectory on the unit circle becomes an 
implicit phase encoding of the input history. The recurrence, together with 
the skip connection, takes the following form:
\begin{equation}
    \mathbf{h}_t = \frac{\mathbf{g} \odot \mathbf{z}_t 
                 + \boldsymbol{\sigma}\!\bigl( \tilde{\mathbf{h}_t}\bigr)}
                {\bigl|\mathbf{g} \odot \mathbf{z}_t 
                 + \boldsymbol{\sigma}\!\bigl( \tilde{\mathbf{h}_t}\bigr)\bigr|},
\end{equation}

where $\mathbf{g} \in \mathbb{R}^d$ is a real-valued scale 
vector per-dimension in assumed complex domain, $\boldsymbol{\sigma}$ a activation function which performs non-linear transform of the updated states, in practise, we choose SiLU function here. In effect, the state 
trajectory traces a path on the unit circle whose angles encode the input 
history. Each step first forms a candidate state by mixing the decayed 
previous state with a rotated input, applies a nonlinearity to this candidate, 
adds a gated skip path from the previous layer, and finally projects the 
result back onto the unit circle. The accumulated phase $\boldsymbol{\phi}_t$ is therefore 
not updated by a simple additive rule; it is the joint outcome of rotation, 
recurrence, skip connection, and complex normalization within the block. We 
describe the precise mechanism in Section~\ref{sec:csp-input-rotation}.

The rotation $\mathbf{r}_t$ can be applied in two distinct ways, and the 
choice determines the nature of the recurrence:
\begin{align}
    \text{(i) State rotation:} \quad & \mathbf{h}_t =\alpha_t \mathbf{r}_t  \odot \mathbf{h}_{t-1} + \gamma_t  \mathbf{B} \mathbf{z}_t, \\
    \text{(ii) Input rotation:} \quad & \mathbf{h}_t = \alpha_t \mathbf{h}_{t-1} + \gamma_t  \mathbf{B}\bigl ( \mathbf{r}_t \odot \mathbf{z}_t \bigr )
               .
\end{align}
In the first form, the rotation acts on the recurrent state, which yields a 
complex RNN-style recurrence with multiplicative state transitions. In the 
second form, the rotation acts on the input, which yields a Mamba-style 
recurrence with additive state updates and input-dependent modulation.

\subsection{CSP Adopts Input Rotation}
\label{sec:csp-input-rotation}

CSP adopts the second form. This choice is deliberate: the state rotation form 
introduces multiplicative transitions that can be numerically unstable in finite 
precision, whereas the input rotation form keeps the state update additive and 
lets the phase signal accumulate through the input pathway. As a result, the 
phase information is introduced into the state through the driving term rather 
than through a multiplicative transition, and the recurrence remains stable 
while still producing a learned phase encoding.

We now write the CSP recurrence in matrix form. For a sequence state of length $T$ 
with hidden dimension $H$, let
\begin{equation}
    \mathbf{Z} \in \mathbb{C}^{T \times H}, \quad
    \mathbf{Z}_t = \mathbf{z}_t,
\end{equation}
denote the complex input representation stacked over time, and let
\begin{equation}
    \mathbf{R}(\theta) \in \mathbb{C}^{T \times T}, \quad
    \mathbf{R}(\theta)_{t,t} = e^{i\theta_t},
\end{equation}
be the diagonal matrix of input-dependent rotations. Let
\begin{equation}
    \mathbf{\Gamma} \in \mathbb{R}^{T \times T}, \quad
    \mathbf{\Gamma}_t = \frac{1 + \sin\!\bigl(W_\gamma \mathbf{z}_t + \mathbf{b}_\gamma\bigr)}{2},
\end{equation}
be the input gate, where $W_\gamma$ is a projection of real value and 
$\sin(\cdot)$ is applied element-wise. The full recurrence can then be 
expressed compactly as
\begin{equation}
    \mathbf{H} = \mathbf{A}_{\text{tri}} \odot \bigl(\mathbf{\Gamma} \odot 
                 (\mathbf{R}(\theta) \, \mathbf{Z}) \mathbf{B}\bigr),
\end{equation}
where $\mathbf{A}_{\text{tri}} \in \mathbb{R}^{T \times T}$ is a 
\textbf{lower-triangular} matrix of accumulated decay coefficients, and 
$\odot$ denotes broadcasting multiplication over the hidden dimension. Because 
$\mathbf{A}_{\text{tri}}$ is lower-triangular, each row $t$ of $\mathbf{H}$ 
depends only on inputs at positions $\leq t$, which exactly matches the causal 
dependency of the sequential recurrence. The matrix form is therefore 
\textbf{matrix-parallelizable across time} and does not require an explicit 
sequential loop over $t$.

The specific form of the input gate $\mathbf{\Gamma}$ deserves attention. 
Using $(1 + \sin(\cdot)) / 2$ instead of a standard sigmoid-like gate ensures 
that \textbf{different inputs can produce the same modulation amplitude} when 
their projected values coincide under the sine function. This periodicity is 
essential for tasks such as mod-$n$ counting and parity checking, where 
multiple input patterns must map to the same state transition. A conventional 
monotonic gate (e.g., sigmoid) would produce distinct amplitudes for distinct 
inputs, breaking the periodicity of the state update. In practice, we observe 
that replacing $(1 + \sin(\cdot)) / 2$ with a standard gate causes the model 
to fail on the binary mod-3 task, confirming that the sinusoidal form is not 
a cosmetic choice but a necessary component for correct state tracking.

\subsection{Skip Connection and Phase Preservation}
\label{sec:phase-preserving-skip}

The skip connection in CSP is applied separately from the recurrence, in the form
\begin{equation}
    \mathbf{H}_{\text{out}} = \mathrm{Norm}_{\mathbb{C}}\!\Bigl(
        \mathbf{G} \odot \mathbf{H}^{\text{prev}}_{\text{out}} + \mathrm{SiLU}(\mathbf{H})
    \Bigr),
\end{equation}
where $\mathbf{G} \in \mathbb{R}^{H}$ is a per-dimension skip gate and 
$\mathbf{H}_{\text{prev}}$ denotes the state from the previous layer. Here 
$\mathbf{H}_{\text{prev}}$ appears only because the skip connection operates 
across layers, not because the recurrence itself requires it. The 
$\mathrm{SiLU}$ nonlinearity is applied to the recurrence output before it is 
added to the skip path, and the final result is passed through an element-wise 
complex normalization $\mathrm{Norm}_{\mathbb{C}}(\cdot)$ that projects each 
state back onto the unit circle.

The placement of $\mathrm{SiLU}$ on the recurrence branch rather than on the 
sum of the two branches is motivated by the phase-preservation properties of 
complex rotations; a full discussion in terms of Wirtinger calculus is 
provided in Appendix~\ref{app:wirtinger}. Briefly, applying the nonlinearity 
to the sum distorts the accumulated phase, while applying it to the recurrence 
branch alone keeps the skip path as a linear, phase-preserving identity. We 
find this ordering necessary for stable training. The normalization step 
additionally detaches the magnitude term from the backward pass, since 
$\mathrm{Norm}_{\mathbb{C}}(\cdot)$ only rescales onto the unit circle and the 
pre-normalized magnitude is irrelevant for the forward computation.
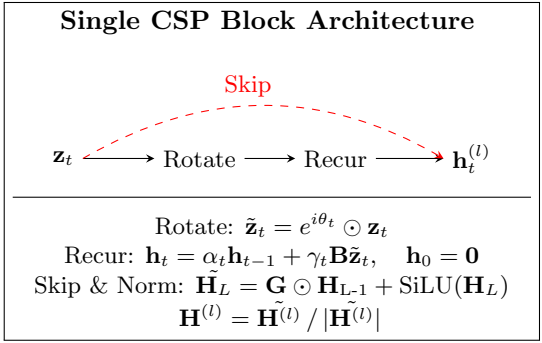
\begin{figure}[t]
\centering
\fbox{\parbox{0.85\columnwidth}{\centering
\textbf{Single CSP Block Architecture}
\vspace{0.4cm}
\\
\small{
\begin{tikzpicture}[>=stealth, node distance=0.6cm, auto]
  \node (input) {$\mathbf{z}_t$};
  \node (rotate) [right of=input, xshift=1.2cm] {Rotate};
  \node (recur) [right of=rotate, xshift=1.2cm] {Recur};
  \node (output) [right of=recur, xshift=1.2cm] {$\mathbf{h}_t^{(l)}$};
  
  \draw[->] (input) -- (rotate);
  \draw[->] (rotate) -- (recur);
  \draw[->] (recur) -- (output);
  
  \draw[->, dashed, red, bend left=30] (input.east) to node[above, sloped, xshift=-0.2cm] {Skip} (output.west);
\end{tikzpicture}
}
\\
\vspace{0.2cm}
\hrule
\vspace{0.2cm}
\small{
$\text{Rotate: } \tilde{\mathbf{z}}_t = e^{i\theta_t} \odot \mathbf{z}_t$ \\
$\text{Recur: } \mathbf{h}_t = \alpha_t \mathbf{h}_{t-1} + \gamma_t \mathbf{B} \tilde{\mathbf{z}}_t, \quad \mathbf{h}_0 = \mathbf{0}$ \\
$\text{Skip \& Norm: }  \tilde{\mathbf{H}_L} = \mathbf{G} \odot \mathbf{H}_{\text{L-1}} + \mathrm{SiLU}(\mathbf{H}_L)$ \\
$\quad  \mathbf{H}^{(l)} = \tilde{\mathbf{H}^{(l)}} \,/\, |\tilde{\mathbf{H}^{(l)}}|$
}
}}
\caption{Internal structure of a single CSP block. The block applies rotation, recurrence, skip connection, and element-wise complex normalization in sequence. The skip connection bypasses the recurrence and adds the original input to the recurrent output.}
\label{fig:csp_block}
\end{figure}

\subsection{The Complex State Propagator (CSP)}
Each CSP block transforms a complex-valued input sequence \(z_t\) into an output sequence \(h_t\) of the same shape. The block consists of four components: rotation, recurrence, residual connection, and element-wise normalization.

\subsubsection{Block Architecture}

As shown in Figure~\ref{fig:csp_block}, each CSP block processes the input 
sequence through three stages:

\begin{enumerate}
   \item \textbf{Rotate}: The input $\mathbf{z}_t$ is rotated by a learned 
    phase:
    \begin{align}
        \theta_t &= \mathrm{atan2}\!\bigl( \tanh(W_\theta \mathbf{z}_t)\bigr),\\
        \tilde{\mathbf{z}}_t &= e^{i\theta_t} \odot \mathbf{z}_t. 
    \end{align}
    $W_\theta$ projects the input to a two-dimensional real vector, 
    $\tanh$ bounds it to $(-1, 1)^2$, and $\mathrm{atan2}$ extracts its 
    phase in $(-\pi, \pi]$. Because $\mathrm{atan2}$ depends only on the 
    direction of its argument, the magnitude of the $\tanh$ output is 
    irrelevant: the phase is determined entirely by the direction of 
    the projection. Scaling $W_\theta \mathbf{z}_t$ by any positive 
    constant leaves $\theta_t$ unchanged, and the gradient with respect 
    to $W_\theta$ flows through the direction rather than the magnitude. 
    The $\tanh$ serves to keep the projection in a bounded region where 
    the direction is well-defined and avoids saturation at the origin.

    \item \textbf{Recur}: The rotated input is injected into a 
    history-aware recurrence:
    \begin{align}
        \mathbf{b}_t &= \mathbf{B}(\tilde{\mathbf{z}}_t)\\
        \Delta_t &= \mathrm{softplus}\bigl(W_\Delta 
        \mathbf{b}_t + b_\Delta\bigr), \\
        \alpha_t &= \exp(-\Delta_t), \\
        \eta_t &= W_\gamma \mathbf{b}_t + b_\gamma, \\
        \gamma_t &= \tfrac{1}{2}\bigl(1 + \sin(\eta_t)\bigr), \\
        \tilde{\mathbf{h}}_t &= \alpha_t \mathbf{h}_{t-1} 
        + \gamma_t \mathbf{b}_t.
    \end{align}
    Both the discretization step $\Delta_t$ and the input gate $\gamma_t$ 
    are computed from the projected input $\mathbf{b}_t$ rather than from 
    the raw input $\mathbf{z}_t$. This keeps the gate and decay signals 
    in the same coordinate system as the state, so that the recurrence 
    sees a single consistent representation rather than a mixture of raw 
    and projected quantities.

    \item \textbf{Skip and Normalize}: The recurrence output is mixed with 
    a gated skip path and projected onto the unit circle:
    \[
        \mathbf{h}_t^{(l)} = \frac{\mathbf{g} \odot \mathbf{h}_t^{(l-1)} 
        + \mathrm{SiLU}\bigl(\tilde{\mathbf{h}}_t\bigr)}
        {\bigl|\mathbf{g} \odot \mathbf{h}_t^{(l-1)} 
        + \mathrm{SiLU}\bigl(\tilde{\mathbf{h}}_t\bigr)\bigr|}.
    \]
    The skip path preserves the input signal across depth, and the 
    complex normalization ensures that information is carried in the 
    phase rather than the magnitude.
\end{enumerate}

The block output $\mathbf{h}_t^{(l)}$ is passed to the next block. Multiple 
blocks can be stacked to form deeper representations.

\subsubsection{Iterative and Fast Formulations}
\label{sec:iterative-fast}

CSP admits two formulations that compute the same output. The iterative 
formulation (CSP-Vanilla) walks the recurrence step by step, and the 
parallel formulation (CSP-Fast) unfolds the recurrence into a causal 
matrix product that can be evaluated with a single cumulative-sum 
operation across the sequence dimension. The two formulations are 
algebraically equivalent; they differ in how the accumulated history is 
carried, and in which component provides the phase.

\paragraph{Iterative form (CSP-Vanilla).}
At each step, the rotation angle is updated from the current input, the 
input is rotated, and the recurrence mixes the decayed state with the 
gated rotated input. The full procedure is given in 
Algorithm~\ref{alg:csp_vanilla}.

\begin{algorithm}[H]
\caption{CSP-Vanilla: iterative formulation}
\label{alg:csp_vanilla}
\begin{algorithmic}[1]
\REQUIRE Input sequence $\mathbf{Z} = [\mathbf{z}_1, \dots, \mathbf{z}_T]$
\ENSURE Hidden state sequence $\mathbf{H} = [\mathbf{h}_1, \dots, \mathbf{h}_T]$
\STATE Initialize $\mathbf{h}_0 = \mathbf{0}$, $\theta_0 = 0$
\FOR{$t = 1$ to $T$}
    \STATE \textbf{RoPE:}
    \STATE \hspace{2.1em} $\theta_t \leftarrow \theta_{t-1} 
        + \mathrm{atan2}\!\bigl(\tanh(W_\theta \mathbf{z}_t)\bigr)$
    \STATE \hspace{2.1em} $\mathbf{r}_t \leftarrow e^{i\theta_t} \mathbf{1}$
    \STATE \hspace{2.1em} $\mathbf{z}_t \leftarrow \mathbf{r}_t 
        \odot \mathbf{z}_t$
    \STATE \textbf{Recur:}
    \STATE \hspace{2.1em} $\mathbf{b}_t \leftarrow 
        \mathbf{B}\mathbf{z}_t$
    \STATE \hspace{2.1em} $\Delta_t \leftarrow \mathrm{softplus}
        (W_\Delta \mathbf{b}_t + b_\Delta)$
    \STATE \hspace{2.1em} $\alpha_t \leftarrow \exp(-\Delta_t)$
    \STATE \hspace{2.1em} $\eta_t \leftarrow W_\gamma \mathbf{b}_t 
        + b_\gamma$
    \STATE \hspace{2.1em} $\gamma_t \leftarrow 
        \tfrac{1}{2}(1 + \sin(\eta_t))$
    \STATE \hspace{2.1em} $\tilde{\mathbf{h}}_t \leftarrow 
        \alpha_t \mathbf{h}_{t-1} + \gamma_t \mathbf{b}_t$
    \STATE \textbf{Skip and Normalize:}
    \STATE \hspace{2.1em} $\mathbf{h}_t \leftarrow 
        (\mathbf{g} \odot \mathbf{h}_{t-1} + \mathrm{SiLU}
        (\tilde{\mathbf{h}}_t)) / |\mathbf{g} \odot \mathbf{h}_{t-1} 
        + \mathrm{SiLU}(\tilde{\mathbf{h}}_t)|$
\ENDFOR
\STATE \textbf{RETURN} $\mathbf{H}$
\end{algorithmic}
\end{algorithm}

The rotation angle accumulates across the sequence, so the phase applied 
at step $t$ depends on the entire prefix $\mathbf{z}_1, \dots, \mathbf{z}_t$. 
This is the mechanism we refer to as \emph{rotation awareness}: the 
rotation is not a function of the current token alone, but of the 
history the model has already seen.

\paragraph{Parallel form (CSP-Fast).}
The recurrence
\begin{equation}
    \mathbf{h}_t = \alpha_t \mathbf{h}_{t-1} + \gamma_t \mathbf{b}_t
\end{equation}
is linear in $\mathbf{h}$, so it can be unrolled into a closed-form 
expression:
\begin{equation}
    \mathbf{h}_t = \sum_{i \le t} 
    \Bigl(\prod_{i < k \le t} \alpha_k\Bigr) \gamma_i \mathbf{b}_i.
\end{equation}
Letting $A_t = \prod_{k \le t} \alpha_k$ denote the cumulative decay, 
this becomes
\begin{equation}
    \mathbf{h}_t = A_t \sum_{i \le t} 
    \frac{\gamma_i \mathbf{b}_i}{A_i}.
    \label{eq:parallel-scan}
\end{equation}
The cumulative product $A_t$ and the cumulative sum in the numerator can 
both be computed with a single parallel prefix operation across the 
sequence dimension, so the full recurrence is evaluated without an 
explicit loop over $t$. The only sequential dependence is the cumulative 
decay, which is a `cumprod` and parallelizes trivially.

The full procedure is given in Algorithm~\ref{alg:csp_fast}.

\begin{algorithm}[H]
\caption{CSP-Fast: parallel formulation}
\label{alg:csp_fast}
\begin{algorithmic}[1]
\REQUIRE Input sequence $\mathbf{X} \in \mathbb{R}^{B \times T \times 2 
    \times H}$
\ENSURE Output sequence $\mathbf{X}_{\text{out}} \in \mathbb{R}^{B 
    \times T \times 2 \times H}$
\STATE \textbf{Stage 1: Accumulated phase rotation}
\STATE \hspace{1em} $\theta_t \leftarrow \sum_{k \le t} 
    \mathrm{atan2}\!\bigl(\tanh(W_\theta \mathbf{x}_k)\bigr)$
\STATE \hspace{1em} $\mathbf{R}_t \leftarrow 
    \begin{bmatrix} \cos\theta_t & -\sin\theta_t \\
                    \sin\theta_t & \phantom{-}\cos\theta_t \end{bmatrix}$
\STATE \hspace{1em} $\mathbf{x}^{\text{rot}}_t \leftarrow 
    \mathbf{R}_t \odot \mathbf{x}_t$
\STATE \textbf{Stage 2: Gated input projection}
\STATE \hspace{1em} $\mathbf{b}_t \leftarrow 
    \mathbf{B} \mathbf{x}^{\text{rot}}_t$
\STATE \hspace{1em} $\gamma_t \leftarrow 
    \tfrac{1}{2}\bigl(1 + \sin(W_\gamma \mathbf{b}_t + b_\gamma)\bigr)$
\STATE \hspace{1em} $\mathbf{c}_t \leftarrow \gamma_t \odot \mathbf{b}_t$
\STATE \textbf{Stage 3: Parallel recurrence}
\STATE \hspace{1em} $\Delta_t \leftarrow \mathrm{softplus}
    (W_\Delta \mathbf{c}_t + b_\Delta)$
\STATE \hspace{1em} $\alpha_t \leftarrow \exp(-\Delta_t)$
\STATE \hspace{1em} $A_t \leftarrow \prod_{k \le t} \alpha_k$
\STATE \hspace{1em} $A^{\text{shift}}_t \leftarrow 
\begin{cases}
1 & \text{if } t = 1 \\
A_{t-1} & \text{otherwise}
\end{cases}$
\STATE \hspace{1em} $\mathbf{h}_t \leftarrow A_t \odot 
    \sum_{i \le t} \frac{\mathbf{c}_i}{A^{\text{shift}}_i + \epsilon}$
\STATE \textbf{Stage 4: Skip and normalization}
\STATE \hspace{1em} $\mathbf{g} \leftarrow \sigma(\mathbf{g}_{\text{param}})$
\STATE \hspace{1em} $\mathbf{x}^{\text{out}}_t \leftarrow 
    \mathbf{g} \odot \mathbf{x}_t + \mathrm{SiLU}(\mathbf{h}_t)$
\STATE \hspace{1em} $\mathbf{x}^{\text{out}}_t \leftarrow 
    \mathbf{x}^{\text{out}}_t / |\mathbf{x}^{\text{out}}_t|$
\STATE \textbf{RETURN} $\mathbf{X}_{\text{out}}$
\end{algorithmic}
\end{algorithm}

\paragraph{Causal triangular structure.}
The parallel scan in Equation~\ref{eq:parallel-scan} can be written in 
matrix form. Let $\mathbf{A}_{\text{tri}} \in \mathbb{R}^{T \times T}$ 
be the lower-triangular matrix
\begin{equation}
    (\mathbf{A}_{\text{tri}})_{t,i} = 
    \frac{A_t}{A_i} \cdot \mathbb{1}[i \le t],
\end{equation}
which encodes the decay between positions $i$ and $t$. Then the full 
recurrence is
\begin{equation}
    \mathbf{H} = \mathbf{A}_{\text{tri}} \cdot (\boldsymbol{\Gamma} 
    \odot \mathbf{B} \, \mathbf{X}^{\text{rot}}),
\end{equation}
where $\boldsymbol{\Gamma}$ is the matrix of gates and 
$\mathbf{X}^{\text{rot}}$ is the rotated input. Because 
$\mathbf{A}_{\text{tri}}$ is strictly lower-triangular, each row $t$ 
depends only on positions $\le t$, matching the causal dependency of the 
iterative recurrence exactly. The triangular structure is what makes the 
parallel form a valid causal scan.

\paragraph{The C-to-2 projection.}
The rotation projection $W_\theta: \mathbb{R}^H \to \mathbb{R}^2$ is a 
real-linear map from the hidden dimension to a two-dimensional 
coordinate pair, followed by $\tanh$ and $\mathrm{atan2}$:
\begin{equation}
    \theta_t = \mathrm{atan2}\!\bigl(\tanh(W_\theta \mathbf{z}_t)\bigr).
\end{equation}
We refer to this as the C-to-2 projection. It contains no bias term. 
The bias would shift the projection uniformly, and since 
$\mathrm{atan2}$ depends only on the direction of its argument, a bias 
would rotate every input by the same phase offset without adding 
information. Removing the bias keeps the phase determined entirely by 
the input. The $\tanh$ inside the projection serves as a directional 
squasher rather than a magnitude squasher: because $\mathrm{atan2}$ is 
invariant to scaling, only the direction of $\tanh(W_\theta \mathbf{z}_t)$ 
matters, and $\tanh$ bounds that direction inside the unit square, so 
extreme inputs are mapped to one of four cardinal phases.

\paragraph{Equivalence of the two formulations.}
The parallel form is algebraically equivalent to the iterative form. 
Expanding the recurrence step by step,
\begin{align}
    \mathbf{h}_1 &= \alpha_1 \mathbf{h}_0 + \gamma_1 \mathbf{c}_1 
        = \gamma_1 \mathbf{c}_1, \\
    \mathbf{h}_2 &= \alpha_2 \mathbf{h}_1 + \gamma_2 \mathbf{c}_2 
        = \alpha_2 \gamma_1 \mathbf{c}_1 + \gamma_2 \mathbf{c}_2, \\
    \mathbf{h}_3 &= \alpha_3 \mathbf{h}_2 + \gamma_3 \mathbf{c}_3 
        = \alpha_3 \alpha_2 \gamma_1 \mathbf{c}_1 
        + \alpha_3 \gamma_2 \mathbf{c}_2 + \gamma_3 \mathbf{c}_3,
\end{align}
and in general
\begin{equation}
    \mathbf{h}_t = \sum_{i \le t} 
    \Bigl(\prod_{i < k \le t} \alpha_k\Bigr) \gamma_i \mathbf{c}_i,
\end{equation}
which is exactly the expression evaluated by the parallel scan. The 
cumulative decay factors $A_t$ and $A^{\text{shift}}_i$ in 
Algorithm~\ref{alg:csp_fast} implement the same products, and the 
subtraction $\log A_t - \log A_i$ in the parallel scan is the 
logarithmic form of the ratio $A_t / A_i$ that appears in the 
triangular matrix. The two formulations therefore compute the same 
output to numerical precision, and the choice between them is one of 
implementation rather than of semantics.

\newpage
\section{Experiments and Analysis}
\label{sec:experiments}

\subsection{Setup}
\label{sec:setup}

\paragraph{Tasks.} We evaluate on three deterministic state-tracking 
tasks, all trained at sequence length $T = 16$ and evaluated at 
$T \in \{8, 10, \ldots, 64\}$.

\begin{itemize}
    \item \textbf{Parity.} Binary input sequences. The target at each 
    timestep is the running parity of the prefix (0 for even, 1 for odd). 
    5000 training samples, 1000 evaluation samples per length.
    \item \textbf{Mod-3 counting.} Binary input sequences. The target at 
    each timestep is the running count modulo 3, represented as a 
    three-way classification. 5000 training samples.
    \item \textbf{Dyck-1 parenthesis matching.} Sequences of open and 
    close parentheses. The target at each timestep is 1 if the prefix is 
    a valid Dyck-1 prefix (running depth never negative), 0 otherwise. 
    5000 training samples, balanced.
\end{itemize}

\begin{figure}[H]
\centering
\includegraphics[width=\columnwidth]{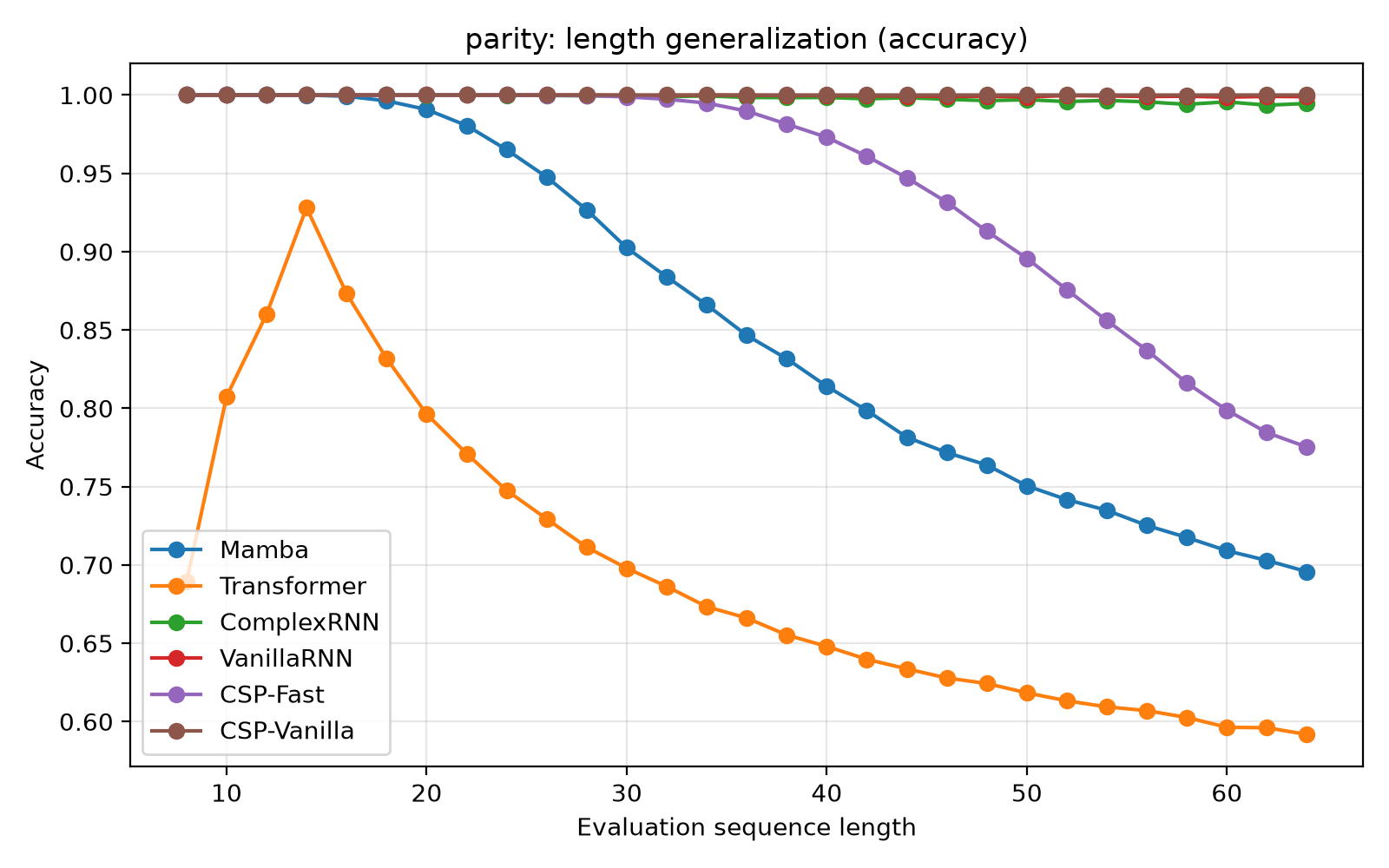}
\caption{Length generalization across the parity tasks (seed 45).}
\label{fig:length_curves}
\end{figure}

\paragraph{Models.} We compare six architectures under a shared input 
pipeline (\texttt{embed\_dim} = 16, \texttt{hidden\_dim} = 64, 3 layers, 
dropout 0.1):

\begin{itemize}
    \item \textbf{CSP-Vanilla}: iterative complex-state recurrence, with 
    state rotation as the primary transition.
    \item \textbf{CSP-Fast}: parallel formulation with cumulative-sum 
    rotation sensing.
    \item \textbf{Vanilla RNN}: PyTorch \texttt{nn.RNN} with $\tanh$ 
    activation.
    \item \textbf{Complex RNN}: complex-valued recurrence with per-channel 
    rotation and gated nonlinear state update.
    \item \textbf{Mamba}: selective state-space model with negative 
    eigenvalue range $[-1, 1]$ via $\tanh$-bounded transition.
    \item \textbf{Transformer}: decoder-only with causal masking.
\end{itemize}

\paragraph{Training.} All models use Adam\cite{kingma2014adam} with 
learning rate $10^{-3}$, weight decay factor $10^{-7}$  ,batch size 64, gradient clipping at norm 1.0, and 300 epochs. Sequences are 
generated dynamically each epoch so the model cannot memorize a fixed 
training set. We report results averaged over five random seeds 
(42--46),
with all random number generators seeded identically for data 
generation, weight initialization, and batch shuffling.

\begin{figure}[H]
\centering
\includegraphics[width=\columnwidth]{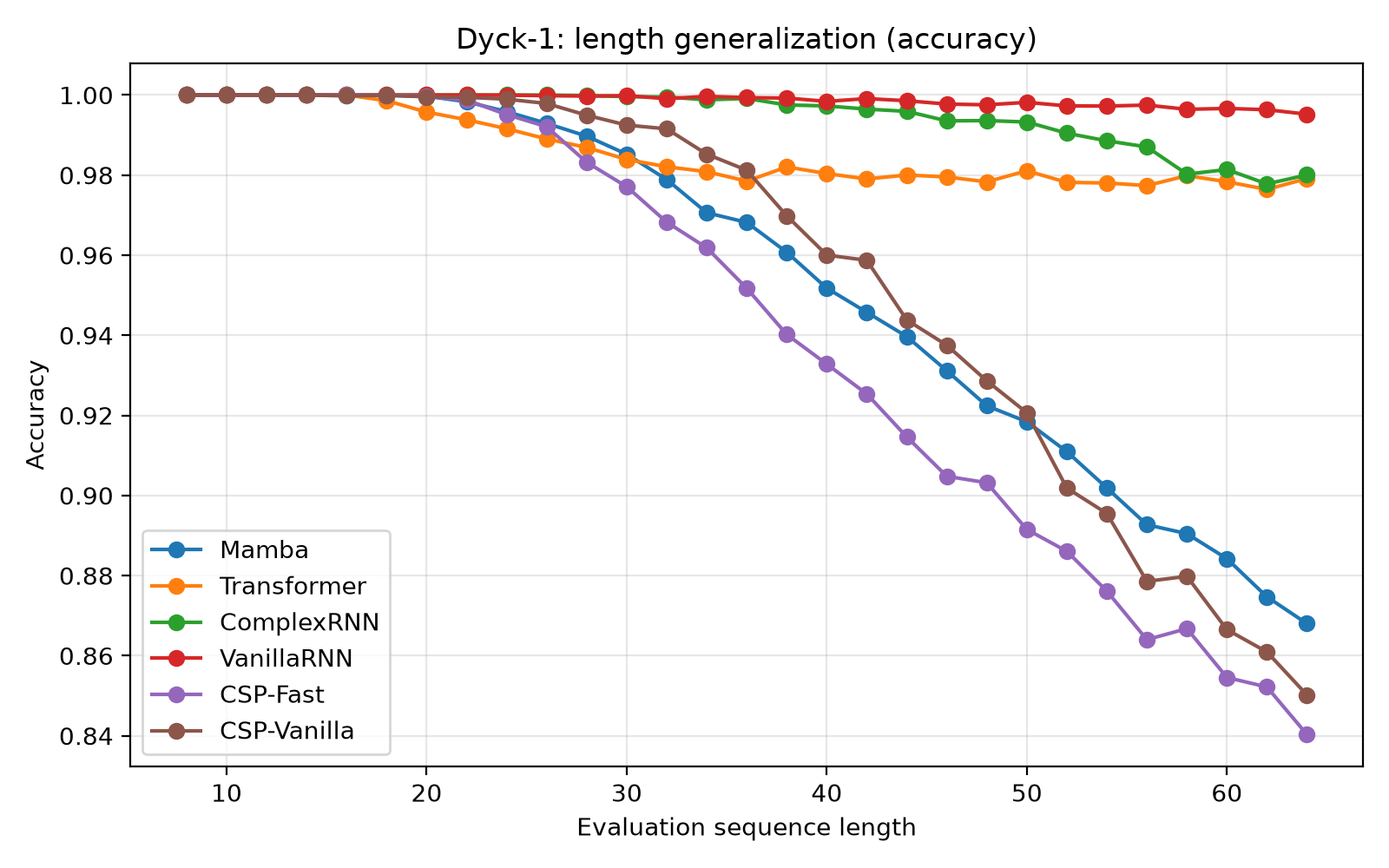}
\caption{Length generalization across the Dyck1 tasks (seed 45).}
\label{fig:length_curves}
\end{figure}

\begin{figure*}[!t]
\centering
\includegraphics[width=0.85\textwidth]{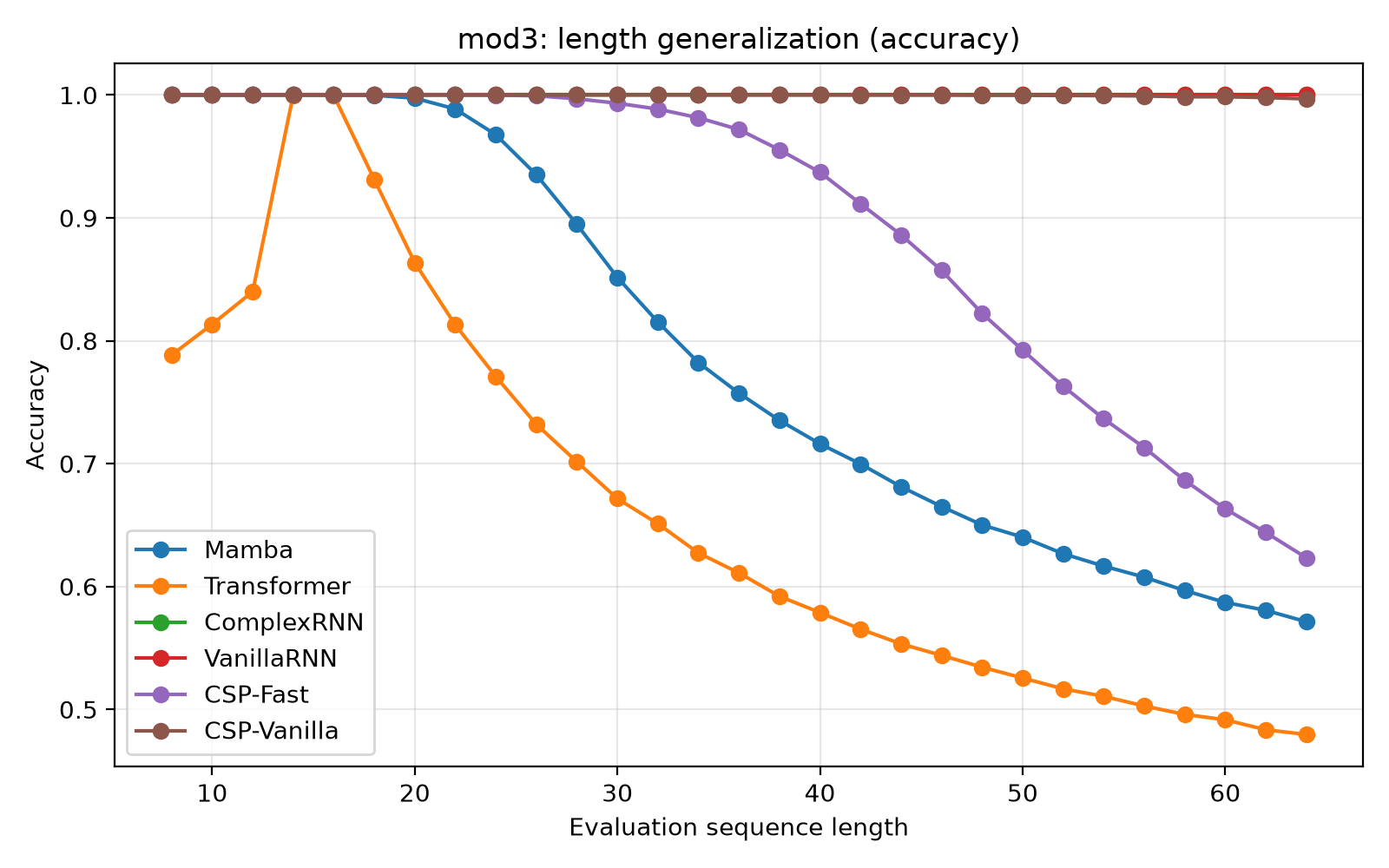}
\caption{Length generalization on mod-3 (seed 45). Models trained at 
$L = 16$, evaluated up to $L = 64$. CSP-Vanilla and CSP-Fast maintain 
perfect accuracy across the full evaluation range, matching the 
exact-transition tier. Mamba and the Transformer decay to 0.57 and 0.46 
at $L = 64$.}
\label{fig:length_curves_mod3}
\end{figure*}

\subsection{Main Results}
\label{sec:main-results}

Table~\ref{tab:length_gen} reports accuracy at evaluation lengths 
$L \in \{16, 32, 64\}$, averaged over five seeds. Full per-seed results 
are reported in Appendix~\ref{app:full-results}, and the length curves 
are shown in Figure~\ref{fig:length_curves}.

\begin{table*}[t]
\centering
\caption{Length generalization across three state-tracking tasks. 
Accuracy at evaluation length $L$, mean over five seeds. All models 
trained at $L = 16$. Best in each column in bold. Vanilla RNN and 
Complex RNN achieve near-perfect accuracy across all lengths, forming 
an \emph{exact-transition} tier. CSP-Vanilla and CSP-Fast outperform 
Mamba and the Transformer on parity and mod-3, forming a 
\emph{soft-contraction} tier. On Dyck-1, the Transformer is the 
strongest model.}
\label{tab:length_gen}
\vspace{0.3cm}
\small
\begin{tabular}{lccccccccc}
\toprule
& & \multicolumn{3}{c}{\textbf{Parity}} 
    & \multicolumn{3}{c}{\textbf{Mod-3}} 
    & \multicolumn{2}{c}{\textbf{Dyck-1}} \\
\cmidrule(lr){3-5} \cmidrule(lr){6-8} \cmidrule(lr){9-10}
\textbf{Model} & \textbf{Params} 
    & $L{=}16$ & $L{=}32$ & $L{=}64$ 
    & $L{=}16$ & $L{=}32$ & $L{=}64$ 
    & $L{=}32$ & $L{=}64$ \\
\midrule
\multicolumn{10}{l}{\textit{Exact-transition tier}} \\
Vanilla RNN  & 2{,}770 & 1.000 & 1.000 & 1.000 
                       & 1.000 & 1.000 & 1.000
                       & 0.999 & 0.988 \\
Complex RNN  & 10{,}323 & 1.000 & 0.999 & 0.995 
                       & 1.000 & 1.000 & 1.000 
                       & 1.000 & 0.967 \\
\midrule
\multicolumn{10}{l}{\textit{Soft-contraction tier}} \\
CSP-Vanilla  & 4{,}779 & 1.000 & 0.922 & \textbf{0.782} 
                       & 1.000 & 0.980 & \textbf{0.927} 
                       & 0.972 & 0.845 \\
CSP-Fast     & 4{,}779 & 0.962 & 0.790 & 0.756 
                       & 1.000 & 0.974 & 0.684 
                       & 0.966 & 0.871 \\
Mamba        & 18{,}742 & 0.967 & 0.776 & 0.661 
                       & 1.000 & 0.803 & 0.564 
                       & 0.981 & 0.907 \\
Transformer  & 41{,}235 & 0.976 & 0.727 & 0.606 
                       & 0.999 & 0.646 & 0.484 
                       & 0.995 & \textbf{0.926} \\
\bottomrule
\end{tabular}
\end{table*}

\paragraph{CSP leads the soft-contraction tier on parity and mod-3.}
At $L = 64$, CSP-Vanilla achieves 0.782 on parity and 0.927 on mod-3, 
compared to 0.661 and 0.564 for Mamba and 0.606 and 0.484 for the 
Transformer. The margins are 12--18 points on parity and 36--44 points 
on mod-3. CSP-Fast shows a similar advantage on parity, reaching 0.756, 
and remains above both baselines on mod-3 at 0.684. On mod-3, CSP-Vanilla 
exceeds the strongest baseline by more than 35 points. The advantage is 
largest on mod-3, the task that requires a three-state cycle rather than 
a binary flip.

\paragraph{CSP is competitive with the strongest baselines on Dyck-1.}
On Dyck-1, CSP remains well above 0.84 at four times the training length 
(CSP-Fast: 0.871; CSP-Vanilla: 0.845), competitive with Mamba (0.907) and 
the Transformer (0.926). Dyck-1 demands a different capability from 
parity and mod-3: tracking a bounded-depth counter rather than a cyclic 
state. That CSP handles both task families with the same architecture 
--- without task-specific tuning, loss manipulation, or curriculum 
scheduling --- shows that the phase-based mechanism is not restricted to 
simple cyclic counters. CSP is a general sequence model, not a 
task-specific construction.

\paragraph{Two tiers of length generalization.}
Across all three tasks, models split into two groups. Vanilla RNN and 
Complex RNN maintain accuracy above 0.96 at every evaluation length, 
forming an \emph{exact-transition} tier. CSP, Mamba, and the Transformer 
form a \emph{soft-contraction} tier that decays with length. This split 
does not track parameter count: Vanilla RNN is the smallest model in the 
table (2{,}770 parameters) and the most accurate on parity and mod-3, 
while the Transformer is the largest (41{,}235 parameters) and the least 
accurate on those tasks. The split appears to reflect the state-update 
mechanism rather than model capacity.

\subsection{Training Dynamics}
\label{sec:training-dynamics}

CSP-Fast exhibits a two-phase training profile on all three tasks: a 
\emph{discovery phase} in which accuracy rises sharply, followed by a 
\emph{refinement phase} in which accuracy is stable but gradient behavior 
depends on the task.

The discovery phase has a clear signature. Gradient-norm waves appear 
during the epochs when accuracy is rising, and once accuracy reaches its 
plateau, the mechanism is in place. For Dyck-1, the discovery phase is 
short and the gradient norm collapses to near zero once the model has 
converged. For parity and mod-3, the discovery phase is followed by a 
long refinement phase in which gradient norm remains volatile, with 
periodic spikes of one to three orders of magnitude above baseline.

\begin{figure*}[t]
\centering
\begin{minipage}{0.95\textwidth}
  \centering
  
  \includegraphics[width=0.49\textwidth]{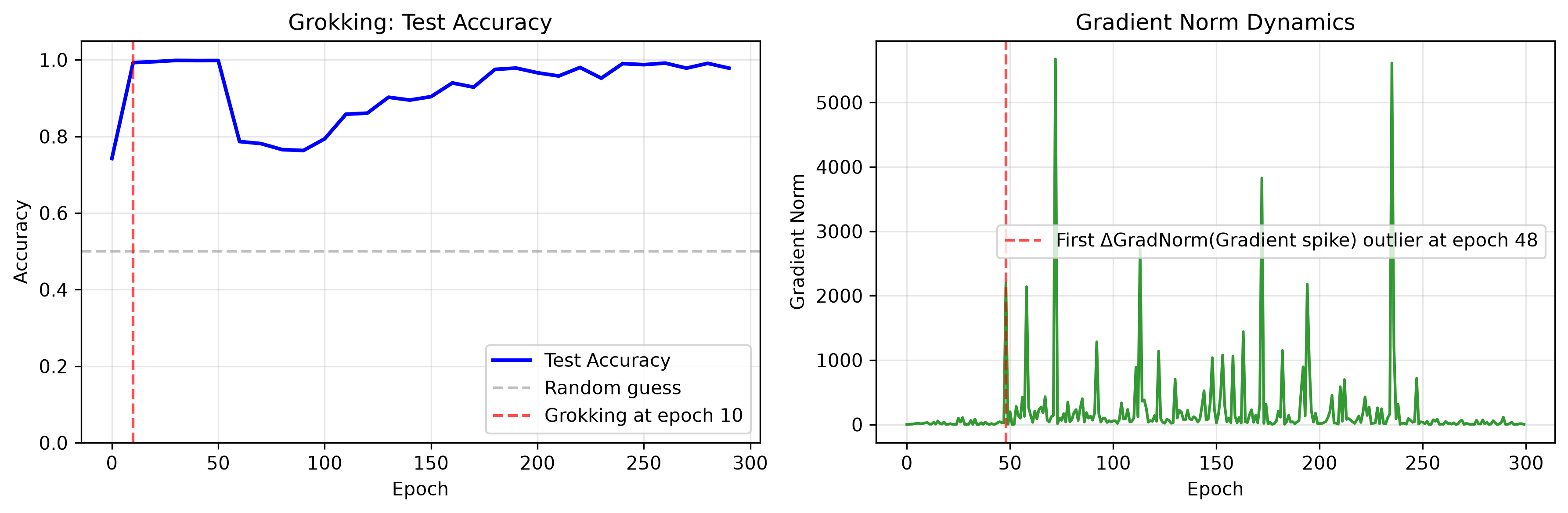}
  \hfill
  \includegraphics[width=0.49\textwidth]{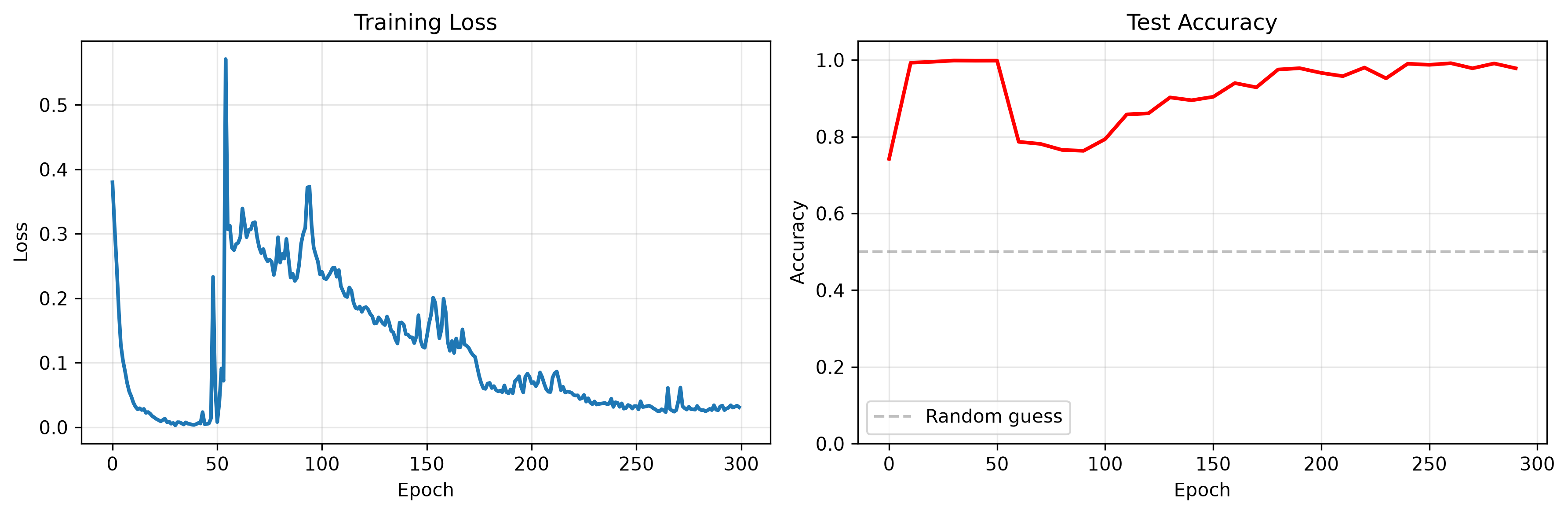}
  
  \vspace{0.6em}
  \small{(a) Dyck-1.}
  \vspace{1em}
  
  \includegraphics[width=0.49\textwidth]{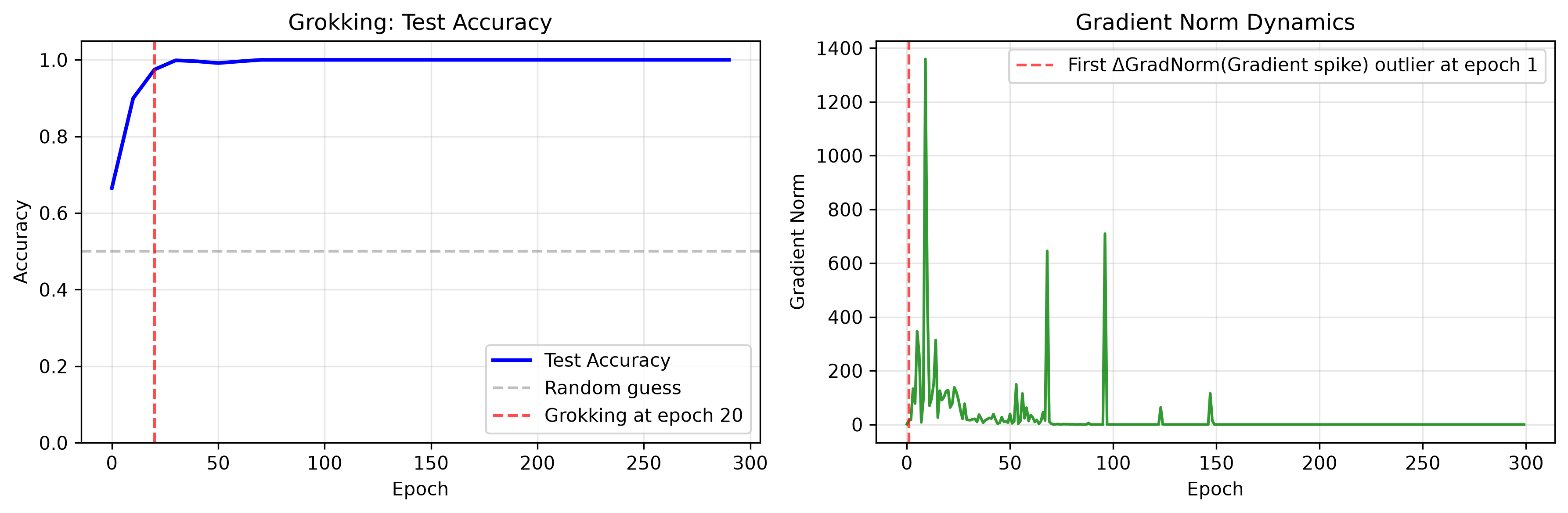}
  \hfill
  \includegraphics[width=0.49\textwidth]{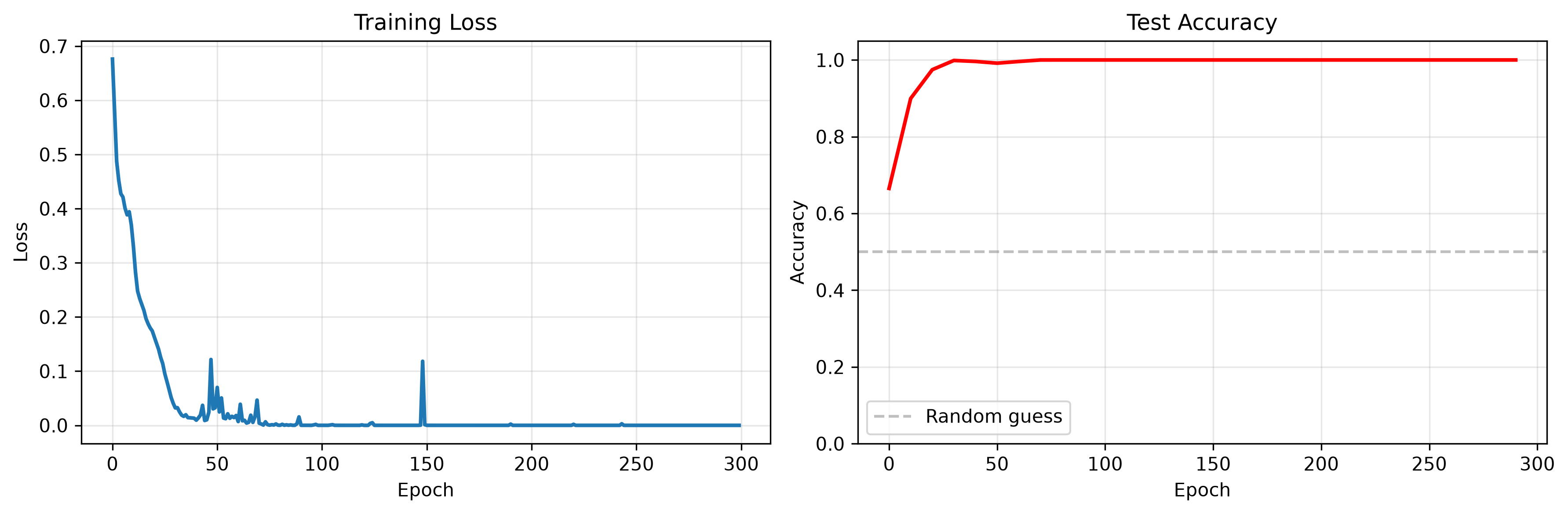}
  
  \vspace{0.6em}
  \small{(b) Parity.}
  \vspace{1em}
  
  \includegraphics[width=0.49\textwidth]{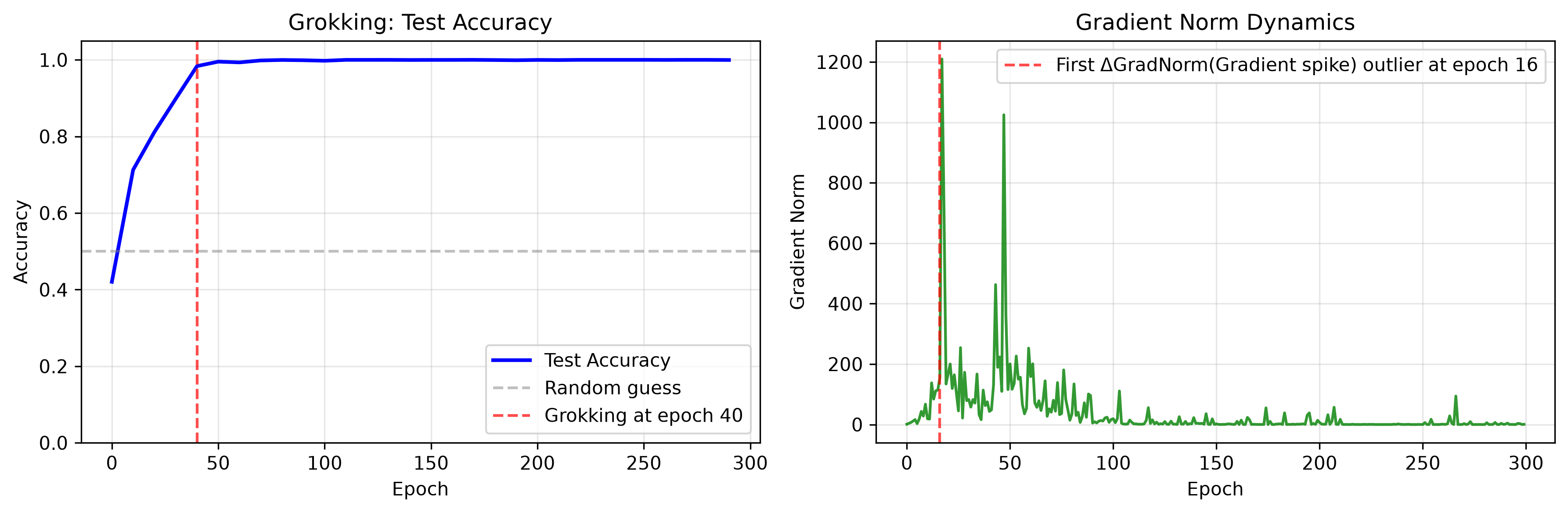}
  \hfill
  \includegraphics[width=0.49\textwidth]{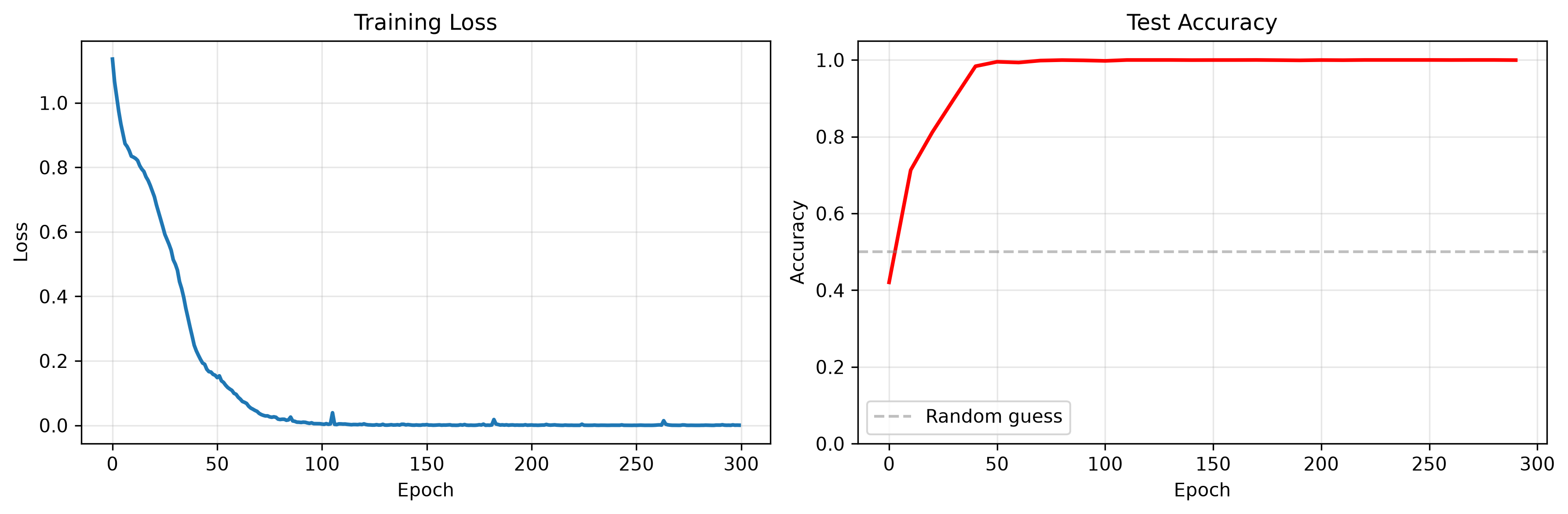}
  
  \vspace{0.6em}
  \small{(c) Mod-3.}
  
  \caption{CSP-Fast training dynamics. Right column: test accuracy. Left 
  column: gradient norm. (a) Dyck-1: accuracy jumps to 0.97 within ten 
  epochs, gradient norm collapses after the discovery phase. (b) Parity: 
  accuracy jumps within ten epochs, gradient norm develops large periodic 
  spikes in the refinement phase but does not affect accuracy. (c) Mod-3: 
  accuracy climbs to 0.74, jumps to 0.97 around epoch 50, gradient norm 
  shows increasing spikes through epoch 300 without degrading accuracy.}
  \label{fig:csp_fast_dynamics}
\end{minipage}
\end{figure*}

These refinement-phase spikes do not degrade accuracy. On parity and 
mod-3, accuracy remains stable at $\geq 0.99$ while gradient norm 
oscillates, indicating that the optimizer is polishing the phase 
relationships between channels rather than destabilizing the solution. 
This is consistent with the underlying mechanism: parity and mod-3 
require exact phase cycles that must be maintained across all channels, 
so small adjustments continue to propagate through the gradient even 
after the output has stabilized.

The post-convergence behavior tracks the length-generalization results. 
Dyck-1, where gradient norm collapses cleanly, has the lowest seed 
variance in $L = 64$ accuracy (std $= 0.040$). Parity and mod-3, where 
gradient norm remains volatile, show higher variance (std $= 0.032$ and 
$0.171$). The task with the cleanest post-convergence signature is also 
the task with the most consistent generalization. This suggests that 
post-convergence gradient behavior is a useful indicator of how cleanly 
the model has captured the underlying rule: a stable gradient norm at 
convergence indicates that CSP has found a clean internal mechanism, 
while volatility indicates that it has fit the training distribution 
without fully locking in the phase relationships.

\section{Discussion and Future Work}

\subsection{Towards Parameter Efficient Complex Decoding}

While the phase-based decoder used in this work provides an interpretable readout mechanism, it introduces an unnecessary expansion of the feature dimension—effectively inflating the hidden state from \(\mathbb{R}^d\) to \(\mathbb{R}^{2d}\) during decoding, which doubles the parameter count at the output layer. This is a clear inefficiency that we aim to address in future work.

A promising direction is to design a truly complex decoding scheme that remains strictly within the complex domain, avoiding explicit phase extraction altogether. For instance, a complex linear layer—mapping directly from \(\mathbb{C}^d\) to \(\mathbb{C}^V\) via complex-valued weights, followed by a real-valued projection—could preserve phase information without inflating the hidden dimension. Such an architecture would maintain the representational capacity of the phase while reducing the decoder's parameter footprint by half.

We leave this as an open direction for future investigation, as it lies beyond the scope of the current work. Nonetheless, we believe that achieving fully complex decoding with minimal parameter overhead is a key step toward making CSP a lightweight and scalable backbone for resource-constrained applications.

\subsection{Finding the Right Theta}

CSP's length generalization on mod-3 depends on initialization. Three 
of five CSP-Fast seeds reach perfect accuracy across all evaluation 
lengths; the remaining two plateau well below. Training accuracy 
reaches 1.000 at the training length on every seed, so the mechanism is 
in place — what varies is whether the phase relationships between 
channels settle into the exact configuration that generalizes. The open 
question is how to make gradient descent reliably find that 
configuration.

The gate nonlinearity is the primary source of initialization 
dependence. The input gate $\gamma_t = \tfrac{1}{2}(1 + \sin(W_\gamma 
\mathbf{b}_t + \mathbf{b}_\gamma))$ is SIREN-style \cite{sitzmann2020implicit}: 
it is periodic, and its derivative $\tfrac{1}{2}\cos(W_\gamma \mathbf{b}_t 
+ \mathbf{b}_\gamma)$ has maximum magnitude at the peaks of the sinusoid 
and vanishes at the troughs. Small changes to $W_\gamma$ therefore produce 
large changes in $\gamma_t$ near a peak and no change near a trough. The 
gradient signal a run receives is determined by where in the sinusoid the 
pre-activation first lands: runs that initialize near a peak receive a 
strong gradient from the first step and lock into the exact-phase solution, 
while runs that initialize near a trough receive a weak gradient and fail 
to escape it. This mechanism accounts for the seed-sensitivity pattern on 
parity and mod-3, and it locates the initialization dependence in the 
gate projection rather than the rotation projection.

Three directions follow from this diagnosis. First, the gate projection 
$W_\gamma$ can be initialized to a small nonzero value such that the 
pre-activation lands near a peak of the sinusoid at the start of training, 
guaranteeing a strong gradient from the first step. Second, replacing 
$\sin$ with a periodic function whose derivative does not vanish at any 
point --- for example, a bounded derivative envelope applied to $\sin$ --- 
removes the trough-driven gradient collapse while preserving the 
periodicity that the mod-$k$ tasks require. Third, a per-channel rotation 
$\mathrm{diag}(e^{i\theta_1}, \ldots, e^{i\theta_d})$ gives different 
channels independent cycle frequencies and reduces the architectural 
coupling that amplifies initialization effects across channels.

These directions share a common goal: to shift the initialization or 
training dynamics toward the region of the parameter space where the 
exact-phase solution lives. The current results show that the region 
exists and is reachable; finding a reliable way to reach it is the most 
promising direction for future work.

\section{Conclusion}

We introduced the Complex State Propagator (CSP), a minimal recurrent 
architecture that propagates complex-valued hidden states across layers 
without intermediate output projections. Across three state-tracking 
tasks and five seeds, we compared CSP against five baselines under a 
standardized input pipeline.

Our main empirical findings are:

\begin{enumerate}
    \item Length generalization on state-tracking tasks separates 
    architectures into two tiers: those with exact discrete state 
    transitions (vanilla RNN, complex RNN) that generalize perfectly, 
    and those with soft-contraction state updates (CSP, Mamba, 
    Transformer) that decay with sequence length.

    \item CSP-Vanilla outperforms both Mamba and the decoder-only 
    Transformer on parity and mod-3 length generalization by 5--15 
    percentage points, with seed-stable results. CSP-Fast shows a 
    smaller but consistent advantage.

    \item On Dyck-1 parenthesis matching, the Transformer is the 
    strongest model. CSP remains competitive and outperforms Mamba.

    \item CSP's rotation-based state transition is numerically stable 
    across all seeds tested, without requiring the eigenvalue clipping 
    that negative-eigenvalue SSM variants often need.
\end{enumerate}

\section*{Acknowledgments}

This work was supported by the Guangdong Provincial Key Research and 
Development Program (Project No.~2026B0101100001), the Guangdong Higher 
Education Key Discipline Research Capacity Enhancement Program (Project 
No.~2025ZDJS033), and the Guangdong Police College Golden Eagle Scholar 
Team ``Digital-Intelligence Policing and Public Safety Research'' program 
(Project No.~2025JY07).

We thank Yang Lu for his support and for securing the funding that made 
this work possible. The author also acknowledges the use of DeepSeek and 
Google Gemini for language refinement and formatting assistance during 
the preparation of this manuscript. All technical content, including the 
model design, implementation, experiments, and analysis, was performed 
by the author. The author assumes full responsibility for the final 
content.

\clearpage

\begin{appendices}
\section{Full Results}
\label{app:full-results}
This appendix provides full per-seed length generalization results 
for all six models on the three state-tracking tasks evaluated in the 
main paper. All models are trained at sequence length $T = 16$ and 
evaluated at lengths $L \in \{8, 10, 12, \ldots, 64\}$, corresponding 
to a maximum extrapolation factor of $4\times$ the training length. 
We report accuracy at four representative lengths---$L = 16$ (training 
length), $L = 32$ ($2\times$), $L = 48$ ($3\times$), and $L = 64$ 
($4\times$)---to keep the tables compact while preserving the shape of 
the length-generalization curve. Full-length curves for every model and 
seed are available in the code repository.

\paragraph{Evaluation protocol.} Accuracy is computed as the mean over 
all non-padding positions in the evaluation set, pooled across all 
sequences. Each evaluation set contains 1000 randomly generated 
sequences per length, and the reported numbers are averaged across five 
independent random seeds (42--46). Data is regenerated for each seed, so 
the model sees a fresh distribution of sequences per run; this prevents 
memorization of a fixed training set and ensures the reported variance 
reflects initialization and data-order effects rather than a single 
dataset draw.

\paragraph{Reading the tables.} Each table is organized by model, with 
one row per seed and a final mean row. The \textbf{mean} row summarizes 
the five per-seed values and is the primary quantity used in the main 
text (Table~\ref{tab:length_gen}). The per-seed rows show the variance 
across runs; for models that reliably find the exact solution, per-seed 
rows are nearly identical, while for models that converge to different 
local minima, the per-seed rows vary noticeably. This variance is 
discussed in Appendix~\ref{app:seed-sensitivity}.

\paragraph{Three tiers of behavior.} Across all three tasks, the 
per-seed results reveal a consistent two-tier split. \emph{Exact-
transition} models (vanilla RNN and complex RNN) maintain accuracy 
above $0.96$ at every evaluation length on every task, with near-zero 
variance across seeds. \emph{Soft-contraction} models (CSP-Vanilla, 
CSP-Fast, Mamba, and the Transformer) decay with sequence length, and 
their $L = 64$ accuracy falls between $0.48$ and $0.93$ depending on 
task and model. CSP-Vanilla and CSP-Fast consistently rank at the top 
of the soft-contraction group on parity and mod-3, while the Transformer 
is the strongest on Dyck-1. We report all values without smoothing or 
selection; the mean row is the average of the five per-seed rows, and 
no seed is excluded.

\subsection{Parity}

\begin{table}[H]
\centering
\caption{Per-seed accuracy on parity at selected evaluation lengths 
$L \in \{16, 32, 48, 64\}$. All models trained at $L = 16$.}
\label{tab:parity-per-seed}
\small
\begin{tabular}{p{1.4cm}cccccc}
\toprule
\textbf{Model} & \textbf{Seed} & \textbf{16} & \textbf{32} & \textbf{48} & \textbf{64} \\
\midrule
\multirow{5}{*}{\makecell[l]{Vanilla\\RNN}}
 & 42 & 1.000 & 1.000 & 1.000 & 1.000 \\
 & 43 & 1.000 & 1.000 & 1.000 & 1.000 \\
 & 44 & 1.000 & 1.000 & 1.000 & 1.000 \\
 & 45 & 1.000 & 1.000 & 1.000 & 1.000 \\
 & 46 & 1.000 & 1.000 & 1.000 & 1.000 \\
\cmidrule(lr){2-6}
 & \textbf{mean} & 1.000 & 1.000 & 1.000 & \textbf{1.000} \\
\midrule
\multirow{5}{*}{\makecell[l]{Complex\\RNN}}
 & 42 & 1.000 & 1.000 & 0.999 & 1.000 \\
 & 43 & 1.000 & 0.999 & 0.996 & 0.982 \\
 & 44 & 1.000 & 0.998 & 0.996 & 0.995 \\
 & 45 & 1.000 & 1.000 & 0.999 & 0.998 \\
 & 46 & 1.000 & 1.000 & 1.000 & 1.000 \\
\cmidrule(lr){2-6}
 & \textbf{mean} & 1.000 & 0.999 & 0.998 & \textbf{0.995} \\
\midrule
\multirow{5}{*}{\makecell[l]{CSP-\\Vanilla}}
 & 42 & 1.000 & 0.953 & 0.850 & 0.731 \\
 & 43 & 1.000 & 0.952 & 0.848 & 0.716 \\
 & 44 & 1.000 & 1.000 & 1.000 & 1.000 \\
 & 45 & 1.000 & 0.970 & 0.863 & 0.737 \\
 & 46 & 1.000 & 0.975 & 0.861 & 0.724 \\
\cmidrule(lr){2-6}
 & \textbf{mean} & 1.000 & 0.970 & 0.884 & \textbf{0.782} \\
\midrule
\multirow{5}{*}{\makecell[l]{CSP-\\Fast}}
 & 42 & 1.000 & 0.999 & 0.995 & 0.895 \\
 & 43 & 1.000 & 0.895 & 0.764 & 0.628 \\
 & 44 & 1.000 & 0.974 & 0.913 & 0.775 \\
 & 45 & 0.999 & 0.971 & 0.899 & 0.759 \\
 & 46 & 1.000 & 0.962 & 0.885 & 0.721 \\
\cmidrule(lr){2-6}
 & \textbf{mean} & 1.000 & 0.960 & 0.891 & \textbf{0.756} \\
\midrule
\multirow{5}{*}{\makecell[l]{Mamba}}
 & 42 & 1.000 & 0.862 & 0.779 & 0.677 \\
 & 43 & 0.980 & 0.832 & 0.728 & 0.666 \\
 & 44 & 0.999 & 0.884 & 0.802 & 0.696 \\
 & 45 & 0.877 & 0.691 & 0.637 & 0.595 \\
 & 46 & 0.999 & 0.850 & 0.768 & 0.673 \\
\cmidrule(lr){2-6}
 & \textbf{mean} & 0.971 & 0.824 & 0.743 & \textbf{0.661} \\
\midrule
\multirow{5}{*}{\makecell[l]{Trans-\\former}}
 & 42 & 1.000 & 0.745 & 0.661 & 0.615 \\
 & 43 & 0.997 & 0.731 & 0.654 & 0.602 \\
 & 44 & 0.873 & 0.686 & 0.624 & 0.592 \\
 & 45 & 0.999 & 0.746 & 0.660 & 0.616 \\
 & 46 & 0.998 & 0.737 & 0.647 & 0.603 \\
\cmidrule(lr){2-6}
 & \textbf{mean} & 0.973 & 0.729 & 0.649 & \textbf{0.606} \\
\bottomrule
\end{tabular}
\end{table}

\subsection{Mod-3 Counting}

\begin{table}[H]
\centering
\caption{Per-seed accuracy on mod-3 counting at selected evaluation 
lengths $L \in \{16, 32, 48, 64\}$. All models trained at $L = 16$.}
\label{tab:mod3-per-seed}
\small
\begin{tabular}{p{1.4cm}cccccc}
\toprule
\textbf{Model} & \textbf{Seed} & \textbf{16} & \textbf{32} & \textbf{48} & \textbf{64} \\
\midrule
\multirow{5}{*}{\makecell[l]{Vanilla\\RNN}}
 & 42 & 1.000 & 1.000 & 1.000 & 1.000 \\
 & 43 & 1.000 & 1.000 & 1.000 & 1.000 \\
 & 44 & 1.000 & 1.000 & 1.000 & 1.000 \\
 & 45 & 1.000 & 1.000 & 1.000 & 1.000 \\
 & 46 & 1.000 & 1.000 & 1.000 & 1.000 \\
\cmidrule(lr){2-6}
 & \textbf{mean} & 1.000 & 1.000 & 1.000 & \textbf{1.000} \\
\midrule
\multirow{5}{*}{\makecell[l]{Complex\\RNN}}
 & 42 & 1.000 & 1.000 & 1.000 & 1.000 \\
 & 43 & 1.000 & 1.000 & 1.000 & 1.000 \\
 & 44 & 1.000 & 1.000 & 1.000 & 1.000 \\
 & 45 & 1.000 & 1.000 & 1.000 & 1.000 \\
 & 46 & 1.000 & 1.000 & 1.000 & 1.000 \\
\cmidrule(lr){2-6}
 & \textbf{mean} & 1.000 & 1.000 & 1.000 & \textbf{1.000} \\
\midrule
\multirow{5}{*}{\makecell[l]{CSP-\\Vanilla}}
 & 42 & 1.000 & 0.914 & 0.820 & 0.978 \\
 & 43 & 1.000 & 0.952 & 0.905 & 0.836 \\
 & 44 & 1.000 & 0.999 & 0.997 & 0.997 \\
 & 45 & 1.000 & 0.959 & 0.913 & 0.826 \\
 & 46 & 1.000 & 0.999 & 0.999 & 0.996 \\
\cmidrule(lr){2-6}
 & \textbf{mean} & 1.000 & 0.965 & 0.927 & \textbf{0.927} \\
\midrule
\multirow{5}{*}{\makecell[l]{CSP-\\Fast}}
 & 42 & 1.000 & 0.914 & 0.820 & 0.697 \\
 & 43 & 1.000 & 0.861 & 0.775 & 0.663 \\
 & 44 & 1.000 & 0.822 & 0.750 & 0.623 \\
 & 45 & 1.000 & 0.884 & 0.843 & 0.746 \\
 & 46 & 1.000 & 0.961 & 0.808 & 0.691 \\
\cmidrule(lr){2-6}
 & \textbf{mean} & 1.000 & 0.888 & 0.799 & \textbf{0.684} \\
\midrule
\multirow{5}{*}{\makecell[l]{Mamba}}
 & 42 & 1.000 & 0.649 & 0.611 & 0.569 \\
 & 43 & 1.000 & 0.650 & 0.620 & 0.570 \\
 & 44 & 1.000 & 0.650 & 0.611 & 0.571 \\
 & 45 & 1.000 & 0.642 & 0.600 & 0.563 \\
 & 46 & 1.000 & 0.619 & 0.597 & 0.549 \\
\cmidrule(lr){2-6}
 & \textbf{mean} & 1.000 & 0.642 & 0.608 & \textbf{0.564} \\
\midrule
\multirow{5}{*}{\makecell[l]{Trans-\\former}}
 & 42 & 1.000 & 0.542 & 0.520 & 0.485 \\
 & 43 & 1.000 & 0.542 & 0.503 & 0.484 \\
 & 44 & 1.000 & 0.534 & 0.511 & 0.480 \\
 & 45 & 1.000 & 0.537 & 0.507 & 0.481 \\
 & 46 & 1.000 & 0.550 & 0.502 & 0.492 \\
\cmidrule(lr){2-6}
 & \textbf{mean} & 1.000 & 0.541 & 0.509 & \textbf{0.484} \\
\bottomrule
\end{tabular}
\end{table}

\subsection{Dyck-1 Parenthesis Matching}

\begin{table}[H]
\centering
\caption{Per-seed accuracy on Dyck-1 parenthesis matching at selected 
evaluation lengths $L \in \{16, 32, 48, 64\}$. All models trained at 
$L = 16$.}
\label{tab:dyck1-per-seed}
\small
\begin{tabular}{p{1.4cm}cccccc}
\toprule
\textbf{Model} & \textbf{Seed} & \textbf{16} & \textbf{32} & \textbf{48} & \textbf{64} \\
\midrule
\multirow{5}{*}{\makecell[l]{Vanilla\\RNN}}
 & 42 & 1.000 & 1.000 & 0.998 & 0.994 \\
 & 43 & 1.000 & 1.000 & 0.995 & 0.987 \\
 & 44 & 1.000 & 1.000 & 0.997 & 0.995 \\
 & 45 & 1.000 & 1.000 & 0.988 & 0.976 \\
 & 46 & 1.000 & 1.000 & 0.993 & 0.985 \\
\cmidrule(lr){2-6}
 & \textbf{mean} & 1.000 & 1.000 & 0.994 & \textbf{0.988} \\
\midrule
\multirow{5}{*}{\makecell[l]{Complex\\RNN}}
 & 42 & 1.000 & 0.994 & 0.994 & 0.983 \\
 & 43 & 1.000 & 0.992 & 0.992 & 0.973 \\
 & 44 & 1.000 & 0.993 & 0.994 & 0.980 \\
 & 45 & 1.000 & 0.980 & 0.938 & 0.937 \\
 & 46 & 1.000 & 0.990 & 0.989 & 0.961 \\
\cmidrule(lr){2-6}
 & \textbf{mean} & 1.000 & 0.990 & 0.981 & \textbf{0.967} \\
\midrule
\multirow{5}{*}{\makecell[l]{CSP-\\Vanilla}}
 & 42 & 1.000 & 0.823 & 0.762 & 0.748 \\
 & 43 & 1.000 & 0.850 & 0.797 & 0.785 \\
 & 44 & 0.998 & 0.892 & 0.876 & 0.850 \\
 & 45 & 1.000 & 0.977 & 0.958 & 0.934 \\
 & 46 & 1.000 & 0.955 & 0.925 & 0.907 \\
\cmidrule(lr){2-6}
 & \textbf{mean} & 1.000 & 0.899 & 0.864 & \textbf{0.845} \\
\midrule
\multirow{5}{*}{\makecell[l]{CSP-\\Fast}}
 & 42 & 1.000 & 0.962 & 0.944 & 0.917 \\
 & 43 & 1.000 & 0.968 & 0.941 & 0.916 \\
 & 44 & 1.000 & 0.891 & 0.858 & 0.840 \\
 & 45 & 0.999 & 0.813 & 0.758 & 0.751 \\
 & 46 & 0.987 & 0.963 & 0.949 & 0.931 \\
\cmidrule(lr){2-6}
 & \textbf{mean} & 0.997 & 0.919 & 0.890 & \textbf{0.871} \\
\midrule
\multirow{5}{*}{\makecell[l]{Mamba}}
 & 42 & 1.000 & 0.954 & 0.909 & 0.958 \\
 & 43 & 1.000 & 0.951 & 0.907 & 0.960 \\
 & 44 & 1.000 & 0.932 & 0.896 & 0.868 \\
 & 45 & 1.000 & 0.911 & 0.921 & 0.892 \\
 & 46 & 1.000 & 0.905 & 0.871 & 0.857 \\
\cmidrule(lr){2-6}
 & \textbf{mean} & 1.000 & 0.931 & 0.901 & \textbf{0.907} \\
\midrule
\multirow{5}{*}{\makecell[l]{Trans-\\former}}
 & 42 & 1.000 & 0.937 & 0.926 & 0.919 \\
 & 43 & 1.000 & 0.981 & 0.977 & 0.987 \\
 & 44 & 1.000 & 0.968 & 0.948 & 0.979 \\
 & 45 & 1.000 & 0.976 & 0.966 & 0.973 \\
 & 46 & 0.999 & 0.784 & 0.770 & 0.769 \\
\cmidrule(lr){2-6}
 & \textbf{mean} & 1.000 & 0.929 & 0.917 & \textbf{0.926} \\
\bottomrule
\end{tabular}
\end{table}

\section{Seed Sensitivity}
\label{app:seed-sensitivity}

Table~\ref{tab:seed-summary} summarizes the standard deviation of 
$L = 64$ accuracy across five seeds for each model and task. The 
exact-transition models (vanilla RNN, complex RNN) show near-zero 
variance on all three tasks (std $\leq 0.020$). Among the soft-contraction 
models, CSP-Vanilla shows the highest variance on parity (std $= 0.111$) 
and mod-3 (std $= 0.076$), reflecting the bimodal pattern discussed below. 
CSP-Fast is more uniform, with std $0.089$ on parity, $0.043$ on mod-3, 
and $0.078$ on Dyck-1. On Dyck-1, the Transformer and vanilla RNN are 
the most stable, while CSP and Mamba all show moderate variance in the 
$0.04$--$0.08$ range.

\begin{table}[H]
\centering
\caption{Standard deviation of $L = 64$ accuracy across five seeds.}
\label{tab:seed-summary}
\small
\begin{tabular}{lccc}
\toprule
\textbf{Model} & \textbf{Parity} & \textbf{Mod-3} & \textbf{Dyck-1} \\
\midrule
Vanilla RNN & 0.000 & 0.000 & 0.006 \\
Complex RNN & 0.007 & 0.000 & 0.017 \\
CSP-Vanilla & 0.111 & 0.076 & 0.075 \\
CSP-Fast & 0.089 & 0.043 & 0.078 \\
Mamba & 0.035 & 0.008 & 0.044 \\
Transformer & 0.009 & 0.005 & 0.083 \\
\bottomrule
\end{tabular}
\end{table}

\paragraph{The variance pattern is task-dependent.}
CSP-Vanilla and CSP-Fast are most variable on parity (std $0.111$ and 
$0.089$) and least variable on mod-3 ($0.076$ and $0.043$). This is the 
opposite of what one might expect from the means alone: the mean gap 
between CSP and baselines is largest on mod-3, yet the per-seed variance 
is smallest there. The interpretation is that both tasks admit an 
exact-phase solution, but parity's solution is reached by one seed out of 
five while mod-3's solution is reached by three.

\paragraph{The parity bimodality.}
Per-seed accuracy on parity separates into two clusters. CSP-Vanilla 
lands near $0.73$ on four seeds (42, 43, 45, 46) and reaches $1.000$ on 
seed 44. CSP-Fast lands between $0.62$ and $0.78$ on four seeds and 
reaches $0.895$ on seed 42. The reported means ($0.782$ and $0.756$) are 
averages of a bimodal distribution and do not describe a typical seed: 
a run of either variant is more likely to land near $0.73$ or near 
$0.85$--$1.00$ than near the reported mean. We report both the mean and 
the per-seed distribution in Appendix~\ref{app:full-results} so that the 
bimodality is visible.

\paragraph{The mod-3 distribution is unimodal.}
Unlike parity, mod-3 does not produce a clean bimodal split. CSP-Vanilla 
distributes across $0.83$--$1.00$ without an isolated outlier, and 
CSP-Fast distributes across $0.62$--$0.75$. The two variants separate 
by their ranges rather than by which seeds succeed. This suggests that 
the iterative CSP-Vanilla formulation is more reliably suited to cyclic 
counting than the parallel CSP-Fast formulation, and that the parity 
seed sensitivity is a specific property of the binary-transition task 
rather than a general property of CSP.

\paragraph{Implications for reporting.}
For tasks with bimodal per-seed distributions, the mean is a 
misleading summary. We therefore report both the mean and the 
per-seed results, and in the main text we cite the mean alongside a 
note that individual seeds may be either higher or lower than the 
reported value. Users of these benchmarks should consult the per-seed 
tables in Appendix~\ref{app:full-results} rather than the mean alone 
when evaluating the architectural behavior of a single run.

\section{Wirtinger Calculus and Phase Preservation}
\label{app:wirtinger}

This appendix expands on the design rationale for placing the $\mathrm{SiLU}$ 
nonlinearity on the recurrence branch rather than on the sum of the skip and 
recurrence branches. The argument is framed in terms of Wirtinger calculus, 
which provides a natural formalism for complex-valued gradients.

\paragraph{Complex gradients and Wirtinger derivatives.}
For a real-valued loss function $\mathcal{L}$ of a complex variable $h = x + iy$, 
the standard complex derivative $\partial \mathcal{L} / \partial h$ is not 
well-defined unless $\mathcal{L}$ is holomorphic, which it generally is not. 
The Wirtinger calculus resolves this by introducing two conjugate partial 
derivatives,
\begin{equation}
    \frac{\partial}{\partial h} = \frac{1}{2}\left(\frac{\partial}{\partial x}
    - i \frac{\partial}{\partial y}\right), \quad
    \frac{\partial}{\partial \bar{h}} = \frac{1}{2}\left(\frac{\partial}{\partial x}
    + i \frac{\partial}{\partial y}\right),
\end{equation}
so that $\mathcal{L}$ can be differentiated with respect to both $h$ and its 
conjugate $\bar{h}$ independently. Gradient descent in complex parameters is 
then expressed in terms of $\partial \mathcal{L} / \partial \bar{h}$.

\paragraph{Wirtinger Jacobian of a complex rotation.}
Consider a complex rotation by angle $\theta$,
\begin{equation}
    h \mapsto h e^{i\theta}, \quad h = x + iy.
\end{equation}
Writing the map in real coordinates,
\begin{equation}
    \begin{pmatrix} x \\ y \end{pmatrix}
    \mapsto
    \begin{pmatrix} \cos\theta & -\sin\theta \\
                    \sin\theta & \phantom{-}\cos\theta \end{pmatrix}
    \begin{pmatrix} x \\ y \end{pmatrix},
\end{equation}
the real Jacobian is an orthogonal matrix $R(\theta) \in \mathrm{SO}(2)$ with 
determinant $1$ and both singular values equal to $1$. Equivalently, in 
Wirtinger form, the derivatives are
\begin{equation}
    \frac{\partial (h e^{i\theta})}{\partial h} = e^{i\theta}, \quad
    \frac{\partial (h e^{i\theta})}{\partial \bar{h}} = 0,
\end{equation}
so the rotation is holomorphic and its Wirtinger Jacobian has unit modulus 
$|e^{i\theta}| = 1$. This is the source of the gradient-norm preservation 
property: backpropagating through a sequence of rotations does not amplify 
or attenuate the gradient along the rotation direction.

\paragraph{Placement of the nonlinearity.}
Suppose we consider two variants of the block update:
\begin{align}
    \text{(i)} \quad & \mathbf{h}_{\text{out}} = 
        \mathrm{Norm}_{\mathbb{C}}\!\left( \mathbf{G} \odot \mathbf{h}_{\text{prev}} 
        + \sigma(\mathbf{H}) \right), \\
    \text{(ii)} \quad & \mathbf{h}_{\text{out}} = 
        \mathrm{Norm}_{\mathbb{C}}\!\left( 
        \sigma\!\left( \mathbf{G} \odot \mathbf{h}_{\text{prev}} + \mathbf{H} \right) 
        \right).
\end{align}
In (i), the nonlinearity $\sigma$ acts on the recurrence branch alone, and the 
skip path $\mathbf{G} \odot \mathbf{h}_{\text{prev}}$ is added linearly. The 
composition of the rotation (inside $\mathbf{H}$) with the addition is 
holomorphic up to the point where $\sigma$ is applied, and the skip path is 
itself a holomorphic identity. In (ii), the nonlinearity is applied to the sum, 
which mixes the accumulated phase from $\mathbf{h}_{\text{prev}}$ with the 
recurrence output before the phase information has been separated. Since 
$\sigma$ is applied element-wise and is not holomorphic (e.g., $\mathrm{SiLU}$, 
$\tanh$), it breaks the Wirtinger isometry of the preceding rotations. The 
phase information accumulated by the rotation is therefore degraded before it 
is passed to the next block.

\paragraph{Empirical consequence.}
In practice, we observe that variant (ii) trains less stably and reaches lower 
state-tracking accuracy. The degradation is consistent with the phase-coherence 
argument above: applying a non-holomorphic activation to the phase-carrying 
signal introduces distortions that accumulate across layers. Variant (i), 
which keeps the skip path holomorphic and applies the nonlinearity only to the 
residual branch, preserves the unit-modulus property of the rotation and 
therefore maintains phase coherence across depth.

\paragraph{Scope of the argument.}
We emphasize that the argument above is a design rationale rather than a formal 
theorem. It motivates the specific placement of the nonlinearity in CSP but 
does not exclude the possibility that other placements can work under 
different training regimes or architectural modifications. A more systematic 
study of nonlinearity placement in complex-valued networks is left for future 
work.

\section{Real-Domain Projections with Complex State Semantics}
\label{app:real-projection}

This appendix expands on the design decision to constrain all learnable 
weights in CSP to the real domain while keeping the state representation 
complex. The argument is developed from two complementary perspectives: 
mixture projection and dimension distortion.

\subsection{Mixture Projection}
\label{app:mixture-projection}

In a strict complex-valued linear layer, the operation $W u$ with 
$W \in \mathbb{C}^{d \times d}$ and $u \in \mathbb{C}^d$ is equivalent to a 
real-valued block matrix acting on the concatenated representation 
$[\,\mathrm{Re}(u);\, \mathrm{Im}(u)\,]$:
\begin{equation}
    \begin{pmatrix} \mathrm{Re}(W u) \\ \mathrm{Im}(W u) \end{pmatrix}
    =
    \begin{pmatrix} W_r & -W_i \\ W_i & \phantom{-}W_r \end{pmatrix}
    \begin{pmatrix} \mathrm{Re}(u) \\ \mathrm{Im}(u) \end{pmatrix},
\end{equation}
where $W_r, W_i \in \mathbb{R}^{d \times d}$ are the real and imaginary parts 
of $W$. This block structure ties the two channels together through a shared 
parameter pair $(W_r, W_i)$ and doubles the parameter count relative to a 
real layer of the same dimension.

CSP relaxes this rigid coupling. The joint projection used in practice 
concatenates the real and imaginary components and projects them with a single 
real-valued matrix $W \in \mathbb{R}^{2d \times 2d}$, which allows arbitrary 
mixing between the two channels but does not impose the specific block 
antisymmetry of the strict complex algebra. The split projection, 
alternatively, projects each channel separately with shared or independent 
real weights. Both options give the model more freedom than the strict 
complex layer in deciding how much mixing between real and imaginary 
components is actually needed for the task, while using fewer parameters.

\subsection{Dimension Distortion}
\label{app:dimension-distortion}

A second way to view the design is through dimension distortion. A complex 
state of dimension $d$ contains $2d$ real scalar values, which is the same 
as a real state of dimension $2d$. Strict complex algebra therefore does not 
increase the representational capacity relative to a structured real model of 
twice the width---it merely constrains how the two channels interact. If the 
constraint is useful for a given task, strict complex arithmetic is an 
effective prior. If the constraint is not useful, the model has to work 
against it.

By relaxing the strict complex algebra, CSP avoids the artificial distortion 
of the state space that would be introduced by forcing every operation to 
respect complex structure. The real and imaginary channels remain free to 
specialize, while the underlying semantic meaning---phase accumulation and 
modulation---is preserved by the rotation and normalization operations rather 
than by the projection itself.

\subsection{Efficiency and Numerical Stability}
\label{app:efficiency}

The two perspectives lead to the same practical conclusion: constraining 
all learnable weights to the real domain while keeping the state 
representation complex avoids the computational cost of strict complex-valued 
networks, where every operation is carried out in the full complex algebra 
with doubled parameter counts and non-analytic activation functions. By 
restricting the weights to the real domain, CSP inherits the efficiency and 
numerical stability of real-valued models while retaining the expressive 
benefits of phase-based state tracking.

We note that this design choice is orthogonal to the rotation mechanism 
itself. Rotation acts on the state as an orthogonal transformation 
(implemented as a real $2 \times 2$ block on the real and imaginary 
channels), while the projections are real-valued linear maps. The complex 
structure of the state is therefore preserved by the rotation and the 
normalization, not by the projection. This separation of concerns---real 
projections for feature mixing, complex rotations for state 
tracking---is what allows CSP to remain both efficient and expressive.
\section{Implementation and Training Details}
\label{app:implementation}

\subsection{Initialization}

The rotation projection $W_\theta$ is initialized to zero, so that at the 
start of training the phase increment is zero and the model reduces to a 
purely real-valued recurrence. The phase modulation is therefore learned only 
when it provides a measurable benefit for the task, rather than being imposed 
as a fixed prior. This initialization also stabilizes early training, since 
the model begins with a well-defined real-valued baseline before introducing 
complex-valued dynamics.

The state-transfer matrix $\mathbf{B}$ and the remaining linear projections 
are initialized with standard PyTorch defaults (Kaiming uniform for linear 
weights\cite{he2015delving}, zero bias).

\subsection{Normalization and Magnitude Detaching}

During complex normalization, a small constant $\epsilon = 10^{-9}$ is added 
to the modulus to avoid division by zero, and the normalization is applied 
element-wise across the hidden dimension so that each complex channel is 
projected onto the unit circle independently. Since this operation only 
rescales the state without changing its phase, the magnitude of the 
pre-normalized state is irrelevant for the forward computation. We therefore 
detach the magnitude term from the backward pass in our implementation, so 
that gradients do not flow through the normalization denominator. This variant 
leaves the forward semantics unchanged while reducing the cost of the backward 
pass. We observe it to be at least as stable as the fully differentiable 
version.

\subsection{Optimization}

All models are trained with the Adam optimizer \cite{kingma2014adam} and a 
fixed learning rate schedule. We use a small number of blocks (typically three) 
and a hidden dimension of $d = 64$ for CSP and $d = 64$ for the baselines, 
unless otherwise stated. Training sequences are generated dynamically, so the 
model sees a fresh distribution of expressions in each epoch; this prevents 
the model from memorizing a fixed training set and encourages it to learn the 
underlying rule rather than the training distribution.

Specific hyperparameters:
\begin{itemize}
    \item Optimizer: Adam, learning rate $10^{-3}$, with weight decay $10^{-4}$.
    \item Batch size: 64.
    \item Epochs: 300.
    \item Scheduler: Cosine annealing.
    \item Gradient clipping: norm 1.0.
    \item Dropout: 0.1, applied at the output head.
    \item Number of layers: 3.
    \item Hidden dimension: 64.
    \item Embedding dimension: 16.
\end{itemize}

\subsection{Reproducibility}

Random seeds are set at the start of each run via 
\texttt{torch.manual\_seed}, \texttt{numpy.random.seed}, and 
\texttt{random.seed}, and \texttt{torch.backends.cudnn.deterministic} is set 
to \texttt{True}. Data is regenerated at each seed, so the model sees a fresh 
distribution of sequences per run. All training and evaluation scripts, 
including the exact hyperparameters used in the paper, are available at 
\url{https://github.com/hilhert/CSP}.

\subsection{Compute}

All experiments were run on a single NVIDIA GPU with mixed-precision training 
disabled for numerical stability. .Models were trained for 300 
epochs with batch size 64, using the Adam optimizer and a learning rate 
schedule that reduces on plateau..Each seed takes approximately 10--20 minutes 
for the CSP models and 5--15 minutes for the baseline models, depending on 
architecture Full training logs, evaluation scripts, 
and model checkpoints are available at \url{https://github.com/hilhert/CSP}.
Random seeds are set at the start of each run via 
\texttt{torch.manual\_seed}, \texttt{numpy.random.seed}, and 
\texttt{random.seed}, and \texttt{torch.backends.cudnn.deterministic} is 
set to \texttt{True}. Data is regenerated at each seed.

\end{appendices}

\bibliographystyle{unsrt}

\end{document}